\pdfoutput=1
\documentclass[11pt,twoside,a4paper,openright]{book}

\usepackage[utf8]{inputenc}
\usepackage{hyperref}       
\usepackage{url}            
\usepackage[english]{babel}
\usepackage{amsmath}
\usepackage[round]{natbib}
\usepackage{appendix}
\usepackage{tikz}
\usepackage{xcolor}
\usepackage{listings}
\usepackage{pifont}
\usepackage{braille}

\usepackage{pdfpages}
\usepackage{fancyhdr}
\usepackage{amsthm}
\usepackage{graphicx}
\usepackage{caption}
\usepackage{adjustbox}
\usepackage[weather]{ifsym}
\usepackage{fontawesome5}

\newtheorem{xdefinition}{Definition}

\renewcommand{\footrulewidth}{0pt}

\usepackage{amsmath}
\usepackage{amssymb}
\usepackage{amsthm}
\usepackage{mathtools}
\usepackage{thmtools}
\usepackage[mathscr]{eucal}
\usepackage{bm}
\usepackage{bbm}

\newcommand\Bu{\bm{u}}
\newcommand\Bv{\bm{v}}

\newcommand\By{\bm{y}}

\newcommand\BG{\bm{G}}

\newcommand\BU{\bm{U}}

\newcommand\BW{\bm{W}}

\newcommand\BY{\bm{Y}}

\newcommand\argmax{\mathop{\mathrm{argmax}\,}}

\renewcommand{\leq}{\leqslant}

\newcommand{\soft}{\mathrm{softmax}}

\def\1{\bm{1}}

\DeclareMathAlphabet{\mathsfit}{\encodingdefault}{\sfdefault}{m}{sl}
\SetMathAlphabet{\mathsfit}{bold}{\encodingdefault}{\sfdefault}{bx}{n}

\renewcommand{\vec}[1]{\mathbf{#1}}
\newcommand{\x}{\vec x}

\newcommand{\y}{{\vec y}}
\newcommand{\z}{{\vec z}}

\newcommand{\X}{\mathcal{X}}
\newcommand{\Y}{\mathcal{Y}}

\newcommand{\norm}[1]{\left\lVert#1\right\rVert}
\newcommand*\diff{\mathop{}\!\mathrm{d}}

\newtheorem{example}{Example}

\newtheorem{hypothesis}{Hypothesis}

\usepackage{mfirstuc}
\usepackage[nogroupskip,order=letter]{glossaries}

\usepackage[automake]{glossaries-extra}
\usepackage{glossary-longbooktabs}
\usepackage{glossaries-extra-stylemods}
\usepackage{glossary-mcols}
\makeglossaries

\newcommand{\newname}[3]{%
\newglossaryentry{#1}
{
    name=#2,
    description={#3}
}%
 \expandafter\newcommand\csname #1\endcsname{\gls{#1}}%
 \expandafter\newcommand\csname #1up\endcsname{\Gls{#1}}%
}

\newname{ml}{ML}{Machine learning}
\newname{ai}{AI}{Artificial intelligence}
\newname{dl}{DL}{Deep learning}
\newname{rl}{RL}{Reinforcement learning}
\newname{ann}{NN}{Artificial neural network}
\newname{nesy}{NeSy}{Neuro-Symbolic}
\newname{llm}{LLM}{Large Language Model}
\newname{llms}{LLMs}{Large Language Models}
\newname{rlhf}{RLHF}{Reinforcement Learning from Human-Feedback}
\newname{lstm}{LSTM}{Long short-term memory}
\newname{rudder}{RUDDER}{Return decomposition for delayed rewards}
\newname{uda}{UDA}{Unsupervised Domain Adaptation}
\newname{da}{DA}{Domain Adaptation}
\newname{iwa}{IWA}{Importance Weighted Aggregation}

\newname{kcmidpos}{\textit{s}}{Mid position of 1D key-chest environment}
\newname{kckeypos}{\textit{k}}{Key position of 1D key-chest environment}
\newname{kcchestpos}{\textit{c}}{Chest position of 1D key-chest environment}
\newname{scoringaction}{\ensuremath{scr}}{Scoring function for reward redistribution for time step}
\newname{rrscore}{\ensuremath{rr\_score}}{Scoring function for reward redistribution for episode}
\newname{actionleft}{\ensuremath{a_{l}}}{Action to move $1$ position to the left in 1D key-chest environment}
\newname{actionright}{\ensuremath{a_{r}}}{Action to move $1$ position to the right in 1D key-chest environment}
\newname{kcnrnd}{\ensuremath{n_{rnd}}}{Number of random features in 1D key-chest environment}
\newname{kcnobs}{\ensuremath{n_{obs}}}{Number of features of observation space in 1D key-chest environment}
\newname{kcnkeys}{\ensuremath{n_{k}}}{Number of \textit{keys} required to open \textit{chest} in 1D key-chest environment}
\newname{kclprob}{\ensuremath{p_l}}{Probability of losing all \textit{keys} in 1D key-chest environment}

\newname{lstmsparse}{LSTMs}{Sparsely-connected {\lstm}}
\newname{lstmfull}{LSTMf}{Fully-connected {\lstm}}

\newname{returndecompf}{\ensuremath{\psi}}{Return decomposition function}
\newname{returns}{\ensuremath{\BG}}{Return values of episodes}
\newname{estreturns}{\ensuremath{\widehat{\BG}}}{Estimated return values of episodes}
\newname{return}{\ensuremath{g}}{Return value of an episode}
\newname{estreturn}{\ensuremath{\widehat{g}}}{Estimated return value of an episode}
\newname{reward}{\ensuremath{r}}{$r_t$ is the reward at time $t$}
\newname{state}{\ensuremath{s}}{$s_t$ is the environment state at time $t$}
\newname{action}{\ensuremath{a}}{$\action_t$ is the action at time $t$}
\newname{bbeta}{\ensuremath{\beta}}{Inverse temperature of $\soft$ function}
\newname{sigmoid}{\ensuremath{\sigma}}{Sigmoid function}
\newname{statepatterns}{\ensuremath{\BU}}{Raw state patterns (before optional mapping)}
\newname{storedpatterns}{\ensuremath{\BY}}{Raw stored patterns (before optional mapping)}
\newname{statepattern}{\ensuremath{\Bu}}{Raw state pattern (before optional mapping)}
\newname{storedpattern}{\ensuremath{\By}}{Raw stored pattern (before optional mapping)}
\newname{mstatef}{\ensuremath{m_{state}}}{Optional transformation function of raw \statepatterns}
\newname{mstoredf}{\ensuremath{m_{stored}}}{Optional transformation function of raw \storedpatterns}
\newname{nobsf}{\ensuremath{K}}{Number of features of state-action pair representation {\sapo}}
\newname{nhistf}{\ensuremath{J}}{Number of features of history representation {\saph}}
\newname{sapo}{\ensuremath{\Bv^o}}{Representation of the state-action pair as vector}
\newname{saph}{\ensuremath{\Bv^h}}{Representation of the history as vector}
\newname{sapc}{\ensuremath{\Bv}}{Raw stored or state pattern as concatenation $[\sapo; \saph]$}
\newname{resetf}{\ensuremath{f_{reset}}}{Reset gate function for resetting the history \saph}
\newname{resetweights}{\ensuremath{\BW }}{Trainable weights of reset gate \resetf}

\title{Parameter Choice and Neuro-Symbolic Approaches for Deep Domain-Invariant Learning}
\author{Marius-Constantin Dinu}

\makeatletter
\hypersetup{
    colorlinks=false,
    linkcolor=blue,
    citecolor=green,
    urlcolor=magenta,
    pdfborder=0 0 0,
    pdftitle={\@title},
    pdfsubject={Doctoral Thesis to obtain the academic degree of Doktor der technischen Wissenschaften in the Doctoral Program Technische Wissenschaften},
    pdfkeywords={machine learning, deep learning, neuro-symbolic systems,
    domain adaptation, large language models, parameter choice methods, artificial intelligence},
    pdfauthor={\@author},
    pdfstartview=FitH
}
\makeatother

\usepackage{etoolbox}
\usepackage{microtype}
\patchcmd{\chapter}{\thispagestyle{plain}}{\thispagestyle{fancy}}{}{}

\begin{document}
\fancyhf{}
\fancyfoot[LE,RO]{\thepage}

\pagenumbering{Roman}
\thispagestyle{empty}
\includepdf[pages={1}]{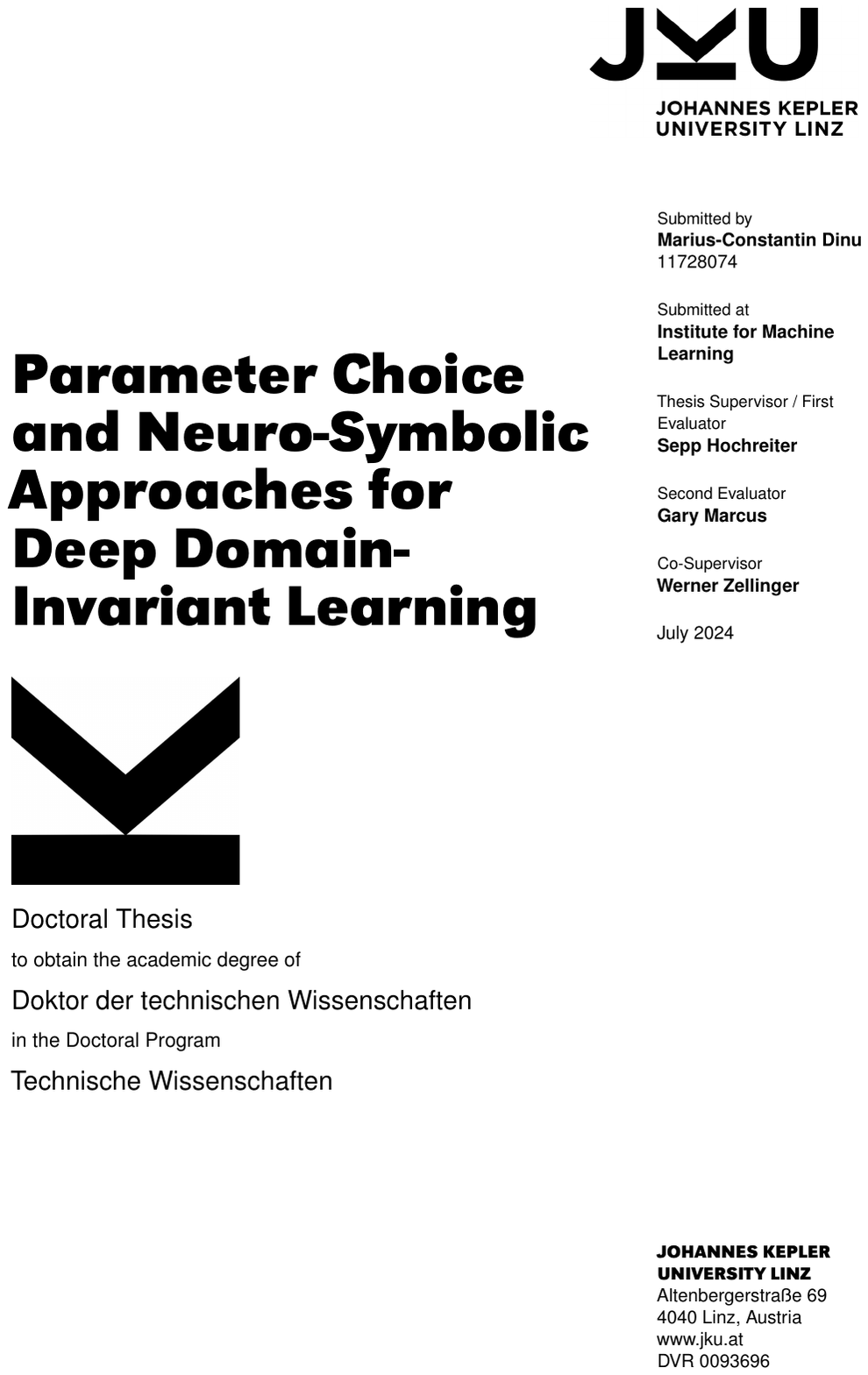}



\chapter*{Acknowledgments}
My academic journey has been supported by countless individuals --- researchers advancing our field, family, colleagues, and friends. Though too numerous to name, their contributions are deeply appreciated.

Firstly, I am deeply indebted to Sepp Hochreiter for establishing this institute, which has significantly shaped my life. His dedication to sharing his knowledge and experience with me has been invaluable, and his expertise continues to inspire me. Words cannot fully capture my appreciation.

Similarly, I express my heartfelt thanks to Werner Zellinger and Sergei Pereverzyev. Werner, in particular, has been a vital mentor during the latter part of my studies. His insightful discussions and advice have been incredibly beneficial. Likewise, Sergei's contributions have been significant, and I've learned a great deal from his work and insights.

I extend my gratitude to my colleagues and friends at the Institute for Machine Learning at Johannes Kepler University. I cherish the engaging discussions, the supportive and friendly environment, the lasting friendships, the exciting conference experiences, and the countless enjoyable moments we've shared. Their support has been critical in navigating a field that is both promising and increasingly challenging, competitive, and sometimes discouraging to pursue a PhD in.
I am grateful to Gary Marcus, whose work and our discussions significantly inspired and contributed to the ideas in this thesis.
I am also thankful to all my co-authors and peers, whom I thoroughly enjoyed collaborating with. Special mentions go to Claudiu Leoveanu-Condrei, Markus Hofmarcher, Markus Holzleitner, Andreas Radler, Kajetan Schweighofer, José Arjona-Medina and Lukas Gruber.
Furthermore, I am grateful to Michael Laux, whose profound wisdom and knowledge, despite brief interactions, offered invaluable guidance and inspiration that will shape my work and perspective.

Finally, I want to express my deepest gratitude to my parents, friends, family, and especially Laura. Your unwavering support has been my cornerstone, without which I could not have pursued my academic goals.

\ \\
\ \\
\ \\
Linz, \today\hfill\textit{Marius-Constantin Dinu}

\newpage
\thispagestyle{empty}

\newpage
\thispagestyle{empty}
\chapter*{Abstract}

As artificial intelligence ({\ai}) systems advance, we move towards broad~{\ai}: systems capable of performing well on diverse tasks, understanding context, and adapting rapidly to new scenarios.
A central challenge for broad~{\ai} systems is to generalize over tasks in related domains and being robust to distribution shifts.
Neuro-symbolic~({\nesy})~{\ai} bridges the gap between symbolic and sub-symbolic paradigms to address these challenges.
This enables us to create adaptable, generalizable, and more interpretable systems.
These characteristics are fundamental for broad~{\ai} systems, which aim to solve a wide range of tasks across various domains.
Consequently, the development of broad~{\ai} requires advancements in domain adaptation ({\da}), enabling models trained on source domains to effectively generalize to unseen target domains.
Traditional approaches often rely on parameter optimization, fine-tuning, and representation learning methods, which can be impractical due to high costs and risks of catastrophic forgetting.
In contrast, {\nesy}~{\ai} systems use multiple models and methods to generalize to unseen domains and maintain performance across varying conditions.
However, building scalable and generalizable hybrid systems remains challenging.

In this work, we analyze common {\da} and {\nesy} approaches with the focus on deep domain-invariant learning.
This analysis extends to real-world challenges, such as adapting to continuously changing domains, handling missing modalities, and dealing with large domain gaps between source and target domains.
We showcase state-of-the-art model-selection and parameter choice methods for scenarios with limited samples where gradient-based optimization is feasible.
For cases where model tuning is infeasible, we introduce domain-specific adaptations without gradient-based updates, connecting in-context learning to domain-invariant learning.
This approach presents {\nesy} methodologies that enable adaptation in previously challenging situations.
Therefore, my work establishes a framework for scalable and generalizable broad~{\ai} systems applicable across various problem settings.
Furthermore, it demonstrates how symbolic reasoning and large language models ({\llms}) can build universal computational graphs that generalize across domains and problems, contributing to more adaptable, generalizable, and interpretable {\ai} approaches for real-world applications.

\includepdf[pages={1-2}]{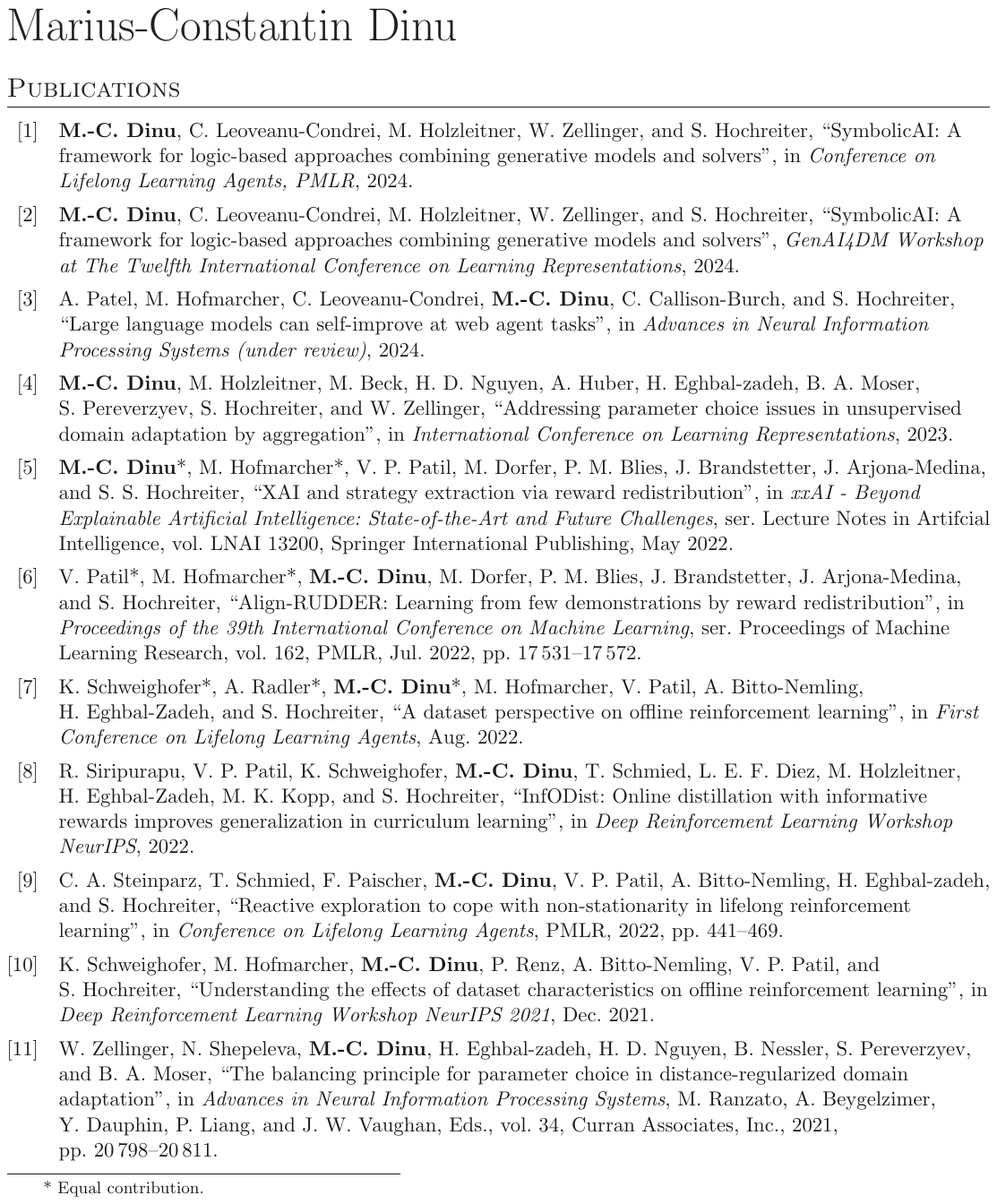}

\newpage
\thispagestyle{empty}

\selectlanguage{english}
\tableofcontents

\newpage
\thispagestyle{empty}

\newpage
\thispagestyle{empty}
\chapter{Introduction}
\fancyhf{}
\fancyfoot[RE]{\leftmark}
\fancyfoot[LO]{\rightmark}
\fancyfoot[LE,RO]{\thepage}
\renewcommand{\footrulewidth}{2pt}
\setcounter{page}{1}
\pagenumbering{arabic}

As artificial intelligence ({\ai}) systems advance, we are moving towards \emph{broad~{\ai}} capable of performing well on diverse tasks, understanding context, and adapting rapidly to new scenarios.
Broad~{\ai}, as conceptualized by \cite{Hochreiter:22}, represents a significant advancement over current narrow {\ai} systems.
It aims to create sophisticated and adaptive systems characterized by enhanced capabilities in knowledge transfer and interaction, adaptability and robustness, abstraction and advanced reasoning, and efficiency.
These systems are designed to quickly adapt to new situations, customers, products, processes, workflows, or sensory inputs, addressing the limitations of current deep learning approaches which often underperform in real-world applications due to their data-hungry nature and limited transfer capabilities.
Key challenges in developing such broad~{\ai} systems include distribution shifts and generalization.
Addressing these challenges is crucial for creating {\ai} systems that can solve a wide range of tasks across various domains and overcome the limitations that currently hinder the widespread adoption of {\ai} in critical real-world scenarios.
Neuro-symbolic~({\nesy})~{\ai} emerges as a promising approach to tackle these challenges by bridging the gap between symbolic and sub-symbolic paradigms of {\ai} \citep{Marcus:01, Marcus:20}. It combines the strengths of both approaches while mitigating their individual weaknesses.
Symbolic~{\ai} excels in logical reasoning, interpretability, and explicit knowledge representation, but struggles with adaptability and learning from raw data.
Sub-symbolic~{\ai}, exemplified by deep neural networks, excels in pattern recognition and learning from large datasets, but often lacks interpretability and struggles with abstract reasoning. {\nesy}~{\ai} aims to create synergy between these approaches by integrating neural learning with symbolic knowledge and reasoning \citep{Garcez:08, Garcez:15, Besold:17, Garcez:19, Garcez:20, Lamb:20, Hamilton:22, Yu:23}.
This integration is challenging due to the fundamentally different nature of these paradigms, which has historically limited their combination.
However, recent advances in machine learning --- specifically due to advances made by large language models ({\llms}) \citep{Vaswani:17, Wei:22CoT, Singhal:23, Schick:23} --- and knowledge representation \citep{Augusto:22} have made it possible to create hybrid systems that leverage the strengths of both approaches.
{\llms} provide a comprehensive natural language understanding and generalization capability, and enable a seamless combination of logical reasoning with statistical approaches.
{\nesy}~{\ai} systems can potentially offer improved adaptability, robustness, generalizability, and interpretability compared to purely symbolic or sub-symbolic systems.
These characteristics are essential for developing broad~{\ai} systems, which are designed to solve a wide range of tasks across various domains and adapt to new challenges without extensive re-training or fine-tuning.
Therefore, to create broad~{\ai} systems, it is essential to advance the field of {\da} and, in particular, domain-invariant learning.

Domain adaptation ({\da}) focuses on training a model on data from a set of source domains to help generalize the model to unseen target domains \citep{widmer1996learning, BenDavid:10, Liu:22}.
{\da}~addresses the domain shift between the source and target domains, which is useful in scenarios where labeled data is scarce or expensive to obtain, making it impractical to re-train models from scratch for each new domain.
Domain-invariant learning is a {\da} methodology that aims to learn shared features that are robust and consistent across different domains.
Core techniques to domain-invariant learning include fine-tuning \citep{Sugiyama07,You19,Zellinger:21},  representation learning  \citep{tzeng2014deep,Ganin16,Sun2017correlation,zellinger2017central,Chen2020moment,Rahman2020,Liu2021kernel,Zhu2021subdomain}, and novel parameter choice methods \citep{Sugiyama07,Sugiyama:12,saito2021tune,Zellinger:21,musgrave2021unsupervised}.
Parameter choice methods involve selecting optimal hyperparameters, regularization techniques, and model architectures to effectively transfer knowledge from source to target domains.
Fine-tuning involves adjusting pre-trained models to adapt to new domains, while representation learning enables the learning of transferable features or embeddings that capture domain-invariant information.
In the context of {\da}, both approaches aim to allow models to adapt to new domains by utilizing shared representations, while minimizing the need for extensive retraining.

However, several problems appear in practice: fine-tuning models that exceed $100$~billion parameters \citep{Brown:20,Google:23,Touvron:23, Jiang:23} is often too computationally expensive or restricted due to proprietary limitations on accessing model weights.
Additionally, the rate of change in real-world data is often intractable.
Therefore, these approaches are not always feasible or applicable.
Furthermore, representation learning approaches lack interpretability and suffer from \emph{catastrophic forgetting} \citep{Hebb1949organization, Carpenter1987, McCloskey1989, mermillod2013stability, ehret2020continual}, when the model is exposed to new domains, which stems from the underlying shift in data distribution across domains.
Catastrophic forgetting refers to a model forgetting previously learned information upon learning new information, particularly in sequential learning tasks.

In contrast to traditional single-model approaches, {\nesy}~{\ai} systems leverage multiple models and methods to address domain-specific requirements and generalize across varying conditions.
They can select between parametric, non-parametric and symbolic methods that incorporate symbolic reasoning up to powerful pattern recognition abilities, structured knowledge bases, and search engines.
This diverse toolkit enables more adaptable and interpretable methodologies.
A hybrid paradigm also enables the formulation of higher-level abstractions and logical expressions through dedicated symbolic approaches.
It incorporates \emph{learning-for-reasoning} methods to treat the learning aspect as an accelerator for reasoning, in which deep neural networks are employed to reduce the search space for symbolic systems \citep{ Silver:16, Silver:17a, Silver:17b, Qu:19, Schrittwieser:20}.
\emph{Reasoning-for-learning} that views reasoning as a way to regularize learning, in which symbolic knowledge acts as a guiding constraint that oversees machine learning tasks \citep{Hu:16, Xu:18}, and \emph{learning-reasoning} to enable a symbiotic relationship between learning and reasoning to boost problem-solving capabilities \citep{Donadello:17, Manhaeve:18, Mao:19, Ellis:23}.
While {\nesy}~{\ai} systems offer numerous advantages, their implementation presents significant challenges.
In practice, building such hybrid systems is challenging due to the inherent complexity of integrating disparate symbolic and neural components, as well as the difficulty in designing interfaces that allow seamless interaction between these components.
This complexity often results in ad-hoc implementations that scale poorly and don't generalize across a variety of tasks and domains.
The lack of standardized frameworks for {\nesy} integration further exacerbates these issues, leading to solutions that are often specific to particular problem domains and difficult to adapt to new scenarios.

Motivated by these limitations and the potential of {\nesy}~{\ai}, my work analyzes common {\da} and {\nesy} approaches, with a particular focus on deep domain-invariant learning.
This analysis is conducted from two complementary perspectives: First, we address {\da} issues faced when dealing with restricted sample sizes and where gradient-based optimization is feasible, and propose novel model-selection and parameter choice methods \citep{Zellinger:21, Dinu:23}.
Second, we establish a connection between in-context learning \citep{Vaswani:17} --- the ability of {\llms} to adapt to new tasks through prompting --- and domain-invariant learning.
This approach focuses on {\nesy} methodologies in situations where tuning models is infeasible, introducing domain-specific adaptations without gradient-based updates \citep{Dinu_SymbolicAI:22, Dinu:24, Dinu:24b}.

As a result, this work summarizes the following contributions:
\begin{itemize}
    \item novel model-selection and parameter choice methods for unsupervised domain adaptation,
    \item a novel {\nesy} approach based on {\llms} in situations where fine-tuning models is infeasible,
    \item a scalable and generalizable framework for broad~{\ai} systems, which enable domain-specific adaptations,
    \item novel benchmarks for parameter choice methodologies and {\nesy} methodologies based on {\llms},
    \item how symbolic reasoning and {\llms} can be connected to build universal computational graphs that generalize across domains and problems.
\end{itemize}
Therefore, our research contributes to the development of more transferable, robust, and interpretable {\ai} approaches that meet the demands of real-world applications, such as robotics, personalized healthcare systems, and business intelligence platforms \citep{TeslaAIDay2022, NVIDIAGTC2024}.

\section{Domain Shift, Domain Adaptation and Domain-Invariant Learning}

The challenges of domain shift and the scarcity of labeled data present significant constraints to traditional deep learning approaches.
To illustrate, in the automotive industry, models designed for classification and segmentation tasks are initially trained in a simulated setting, potentially with full access to labeled data, and aimed at being transferred to real-world applications.
Despite the profusion of unlabeled real-world data, for instance, video recordings under various environmental conditions \citep{TeslaAIDay2022}, this task remains challenging due to intricate domain shifts between the simulated data and the target domain \citep{Dosovitskiy17}.

The domain shift refers to the changes in the data distribution between the source domain (distribution $p$) where the model is trained, and the target domain (distribution $q$) where the model is applied.
This shift can manifest in various forms, such as changes in feature distribution $p_\X \neq q_\X$, label distribution $p_\Y \neq q_\Y$, and conditional distribution between labels given features $p_{\Y \mid \X} \neq q_{\Y \mid \X}$.

To address these challenges systematically, we turn to the field of unsupervised domain adaptation ({\uda}).
{\uda} provides a formal framework for transferring knowledge from a labeled source domain to an unlabeled target domain, directly tackling the issues of domain shift and label scarcity.
In the following section, we present the mathematical foundations of {\uda}, which will serve as the basis for our subsequent analysis and proposed methods.

\paragraph{Unsupervised Domain Adaptation}
Let $\X\subset\mathbb{R}^{d_1}$ be a compact \emph{input space} and $\Y\subset\mathbb{R}^{d_2}$ be a compact \emph{label space}.
Following~\citet{BenDavid:10}, we consider two datasets: A \textit{source dataset} $(\x,\y)=((x_1,y_1),\ldots, \allowbreak (x_n,y_n)) \in \left(\X\times\Y\right)^n$ independently drawn according to some source distribution (probability measure) $p$ on $\X\times\Y$ and an unlabeled \textit{target} dataset $\x'=(x_1',\ldots,x_m')\in\X^m$ with elements independently drawn according to the marginal $q_\X$.
The marginal distribution of $p$ on $\X$ is analogously denoted as $p_{\X}$.
The goal is to find $f: \X \to \Y$ with small \emph{target risk} $\mathcal{R}_q(f)$.
For simplicity, we consider the \textit{expected target risk} $\mathcal{R}_q(f)=\int_{\X\times\Y} \norm{f(x)-y}_{\Y}^2\diff q(x,y)$ with least squares loss.
From this formulation, we can see that {\uda} aims to adapt a model trained on a labeled source domain to perform well on a new, unlabeled target domain, effectively addressing the challenges of domain shift and the absence of labeled data in the target domain \citep{BenDavid:10, Liu:22}.

\begin{figure}[!ht]
    \centering
    \includegraphics[width=1.0\linewidth]{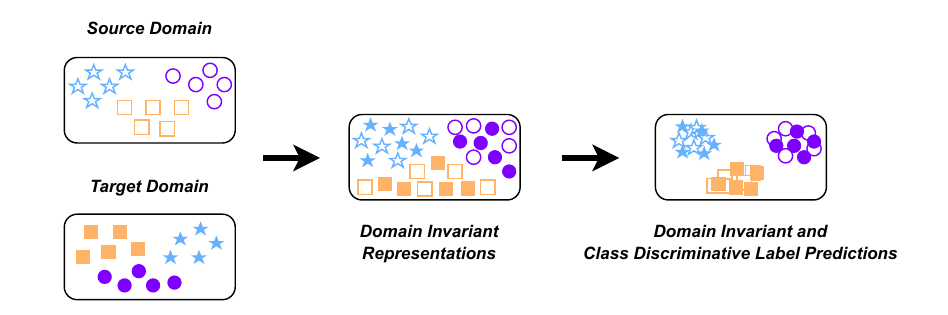}
    \caption{Illustration of domain-invariant learning. Source and target domain features are transformed into domain-invariant features, which are then made class discriminative. This enables the model to distinguish between different classes while maintaining consistent performance across various domains.}
    \label{fig:domain_invariant}
\end{figure}

Domain-invariant learning is a general principle to construct algorithms for {\da} that focuses on learning a feature extractor $\varphi: \X \rightarrow \mathbb{R}^d$ that maps input data $x \in \X$ to a shared $d$-dimensional \emph{representation} space $\mathbb{R}^d$ with $d \geq 1$.
The goal is to obtain representations $\varphi(x)$ that are similar across both the source domain and the target domain.
More specifically, we aim to learn $\varphi$ such that the associated probability distribution of $\varphi(X_p)$ and $\varphi(X_q)$ are as close as possible, where $X_p$ and $X_q$ are random variables with distribution $p$ and $q$, respectively, often measured using distribution distance metrics or divergences such as Maximum Mean Discrepancy \citep{Gretton12}, Wasserstein distance and Central Moment Discrepancy \citep{zellinger2017central}.
See Figure~\ref{fig:domain_invariant} for a visual illustration of domain-invariant learning.

\section{Model Selection and Parameter Choice Methods for Unsupervised Domain Adaptation}
\sectionmark{Model Selection and Parameter Choice Methods}

Implementing effective {\uda} methods requires careful selection of model architectures and hyperparameters.
This section examines approaches to model selection and parameter choice that aim to optimize {\uda} performance across diverse domains, focusing on how these methods can be applied to various {\uda} techniques such as moment matching \citep{zellinger2017central, Gretton12}, divergence-based \citep{Li20}, or discriminative approaches \citep{Ganin16}, among others.

\paragraph{Parameter Choice Problem}
Typical approaches to parameter choice in {\uda} first involve training a sequence of $m$ models with different hyperparameter settings. This often includes modifying parameters such as the learning rate or weighting factors of loss terms.
A current best practice for a second step is then to apply a model selection method to obtain the best-performing model with the minimum risk on the target domain.
However, the state-of-the-art model selection methods \citep{Sugiyama07, You19, Zellinger:21} can only perform as well as the best model within the sequence and disregard the $m-1$ trained models, in spite of the computational power and time invested for training.
Furthermore, the choice of models and parameters often hinges on the task-specific characteristics of the dataset, necessitating dedicated experimentation.

\paragraph{Parameter Choice Contribution} In our work \citep{Dinu:23}, we go beyond selecting a single model from the sequence of models by computing a weighted aggregation of all models within the sequence.

More precisely, we calculate an ensemble $f^{\mathtt{agg}} = \sum_{k=1}^m c_k f_k$  with minimal target risk $\mathcal{R}_q(f^{\mathtt{agg}}) = \min_{c_1, \dots c_m \in \mathbb{R}} \mathcal{R}_q(\sum_{k=1}^m c_k f_k)$.
Here $f_k$ denotes the models within the sequence and $c_k$ the aggregation coefficients.
The optimal aggregation risk is less than or equal to the risk of the best model obtained by model selection, i.e., $\mathcal{R}_q(f^{\mathtt{agg}}) \leq \mathcal{R}_q(f^{\mathtt{sel}})$, because this model is included within the sequence of aggregated models.
Using vector-valued least squares, the vector of optimal coefficients can be calculated as follows (see Section~\ref{sec:addressing}):

\begin{equation} \label{eq:c_coeff}
    c = G^{-1} g = \texttt{arg min}_{(c_1, \dots, c_m) \in \mathbb{R}^m} \int_X \bigg\| \sum_{k=1}^m c_k f_k(x) - f_q(x) \bigg\|_Y^2 \diff q_\mathcal{X}(x),
\end{equation}
where $f_q(x)=\int_Y y\diff q(y|x)$ is the Bayesian predictor (the model with minimal risk on the target domain).
 The Gram matrix $G$ consists of scalar products between the predictions of the individual models within the sequence, i.e. that
 $G = (\int_X  \langle f_k(x), f_u(x) \rangle_Y \diff q_\X(x))_{k, u = 1}^m$,
 and $g = (\int_X \langle f_q(x), f_k(x) \rangle_Y \diff q_\X(x))_{k=1}^m$.
Since calculating $f_q$ requires access to the labels of the target domain, to overcome this problem, under the assumptions of 1) a covariate shift and 2) a bounded density ratio between the marginals of the source distribution $p$ and target distribution $q$, so that $\beta(x) = \frac{\diff q_\X}{\diff p_\X}(x) \in [0, B]$ with some constant $B > 0$, we arrive at the following result:

\begin{equation}
    \label{eq:vec_g}
    g = \bigg(\int_X \langle f_p(x), f_k(x) \rangle_Y \beta(x) \diff p_X(x)\bigg)_{k=1}^m .
\end{equation}

Equation~\eqref{eq:vec_g} computes the expectation over scalar products between the sequence of models and the Bayesian predictor $f_p$ weighted by the density ratio $\beta(x)$ to adjust for the domain shift.

To realize the computations of Equation~\eqref{eq:c_coeff} algorithmically, we employ the following two steps:
First, we estimate the density ratio $\beta(x)$ (see  e.g. \cite{Sugiyama:12} for a collection of possible methods).
Second, we calculate the empirical estimators for the Gram matrix $\widetilde{G}=(\frac{1}{t}\sum_{i=1}^t\langle f_k(x_i') , f_u(x_i') \rangle_{Y})_{k,u=1}^m$ and vector $\widetilde{g}=(\frac{1}{s}\sum_{i=1}^s \beta(x_i) \langle y_i , f_k(x_i) \rangle_{Y})_{k=1}^m$ based on our datasets, and gain aggregation coefficients $\widetilde{c} = \widetilde{G}^{-1}\widetilde{g}$.

Our method defines the state of the art in parameter choice for {\uda}, as it outperforms the current best practices in model selection with theoretical guarantees.
We also performed large-scale empirical evaluations on a text-based domain (AmazonReviews \citep{McAuley21}), an image-based domain (DomainNet \citep{Peng18}), and four time-series datasets \citep{Ragab23}, featuring a heterogeneity dataset for human activity recognition, electroencephalogram recordings of sleep activities, as well as smartphone and smartwatch activity and biometrics datasets.

\section{In-Context Learning}
Large language models ({\llms}) have recently demonstrated promising capabilities \citep{kojima:2022, huang:2022, Ruis:22, Weng:23, Fang:24}. State-of-the-art models typically exceed $10$~billion parameters and exhibit basic reasoning capabilities through in-context learning \citep{Brown:20}, as seen with Chain-of-Thought \citep{Wei:22CoT} and Tree-of-Thought prompting \citep{Yao:23}.

In-context learning is an \emph{emergent phenomenon} in {\llms} that occurs after extensive pretraining on vast datasets. The models have been trained on countless domains and tasks, and their in-context learning capabilities enable them to generalize across various tasks and domains. Unlike traditional learning paradigms that rely on gradient-based updates to adjust model parameters, in-context learning leverages the model's ability to adapt to specific tasks without changing its underlying weights.
We illustrate this with the following example.

\begin{example}
\label{ex:fuzzy_equals}
Prompt-based Instructions:

A typical in-context learning prompt is constructed based on an instruction, e.g.,
"\emph{Perform a fuzzy comparison between the following statements and determine if they are approximately equal or not:}",
some contextual patterns as few-shot examples:
\begin{lstlisting}[language=Python]
1 = "ONE" => True,
6.0 = "SIX" => True,
1 = "two" => False,
"five" = 5 => True,
ten = 10 => True,
4 = "Three" => False,
"four" = "FOUR" => True,
...
\end{lstlisting}
and the new query statement which we want the {\llm} to complete:
\begin{lstlisting}[language=Python]
"eight" = 8 =>
\end{lstlisting}
The {\llm} leverages the provided examples to create associations between numbers and words, allowing it to interpret the \emph{fuzzy comparison} between the presented values and extrapolate to similar examples. This enables the {\llm} to infer that "\texttt{eight}" and "\texttt{8}" are approximately equal in this context, leading it to predict "\texttt{True}" as the correct outcome for the new query statement.
\end{example}

\paragraph{Large Language Models}
To begin our analysis, we first consider a simple prompt: "Hello, my world!", and its possible tokenization:
\begin{equation*}
[\text{"He"}, \text{"l"}, \text{"l"}, \text{"o"}, \text{","}, \text{"my"}, \text{"w"}, \text{"o"}, \text{"r"}, \text{"ld"}, \text{"!"}]
\end{equation*}
The vocabulary $\mathcal{V}$ might then include these tokens and many others:
\begin{equation*}
\mathcal{V} = \{\text{"He"}, \text{"l"}, \text{"o"}, \text{"w"}, \text{"r"}, \text{"ld"}, \text{"!"}, \text{","}, \text{"my"}, \text{"us"}, \text{"of"}, \ldots\}
\end{equation*}
This example illustrates how a sentence is converted into tokens, and how the vocabulary contains whole words, subword units and characters.

More formally, let $\mathcal{V}$ be a finite set representing the vocabulary of the language model. We then define $\mathcal{V}^*$ as the set of all possible token sequences (including the empty sequence) of elements from the finite set of tokens $\mathcal{V}$.
We call an element $x \in \X \subset \mathcal{V}^*$ a \emph{sequences of tokens} from the \emph{input space} and an element $y \in \Y \subset \mathcal{V}^*$ a sequence of tokens from the \emph{output space}.

For our analysis, we first define the subspace $\mathcal{P} \subset \mathcal{X} \subset \mathcal{V}^*$ as the space of all meaningful prompts and subspaces $\mathcal{I} \subset \X$, $\mathcal{C} \subset \X \times \mathcal{Y}$, and $\mathcal{S} \subset \X$ representing the instruction space, context space, and query space, respectively.
We then define the prompt input $x \in \mathcal{P} = \mathcal{I} \times \mathcal{C} \times \mathcal{S}$ as a tuple $x = (i, c, s)$, where $i \in \mathcal{I}$, $c \in \mathcal{C}$, and $s \in \mathcal{S}$.
Now, let $f: \mathcal{P} \to \Y$ represent the {\llm} which predicts outputs with desired targets $y = f(x) \in \Y$.

\paragraph{In-Context Learning and Domain-Invariant Learning}
In the context of domain-invariant learning, we consider the prompt domain $\mathcal{P}$, where samples are drawn from two probability distributions over this domain: a source distribution $p(i, c, s)$ and a target distribution $q(i, c, s)$. The basic intuition is that an {\llm} $f$ learns representations that are invariant across different contexts, enabling generalization from source samples to target samples.
In-context learning then minimizes the expected target risk $\mathcal{R}_q(f)$ not by explicitly updating the model parameters, but through implicit adaptation based on the provided context examples, which is often referred to as \emph{prompt engineering} \citep{Nori:23}.

The in-context learning approach can then be summarized as follows:
\begin{enumerate}
\item \textbf{Encoding:} Given an embedding function $\rho : \mathcal{V} \to \mathbb{R}^{d}$ with $d \geq 1$ we encode the instructions $i \in \mathcal{I}$, context $c \in \mathcal{C}$, and query statement $s \in \mathcal{S}$ into vector representations:
\[
    \mathbf{i}_{1:m} = (\rho(i_1), \rho(i_2), \ldots, \rho(i_m)) \in \mathbb{R}^{m \times d} = (\rho(i_j))^m_{j=1},
\]
\[
    \mathbf{c}_{1:k} = (\rho(c_1), \rho(c_2), \ldots, \rho(c_k)) \in \mathbb{R}^{k \times d} = (\rho(c_j))^k_{j=1},
\]
\[
    \mathbf{s}_{1:n} = (\rho(s_1), \rho(s_2), \ldots, \rho(s_n)) \in \mathbb{R}^{n \times d} = (\rho(s_j))^n_{j=1}.
\]
\item \textbf{Concatenation:} The embedded vectors are concatenated into a single sequence (with lengths $k_1$ and $k_2$ of the two context sequences):
\[
    \mathbf{z}^1_{1:j=m+k_1+k_2+n} = [\mathbf{i}_{1:m}, \mathbf{c}_{1:k_1+k_2}, \mathbf{s}_{1:n}] = (z^1_i)^{j=m+k_1+k_2+n}_{i=1}.
\]
\item \textbf{Transformations:} Transformation functions $\varphi^l_{\mathrm{trans}}$ (e.g., attention mechanisms, feed-forward neural networks, normalization, etc.) are applied across different hidden layers with $1 < l < L$, such that the latent sequences $\mathbf{z}^{l-1}_{1:j}$ produces the subsequent latent representation $\mathbf{z}^{l}_{1:j}$ with latent dimension $d_l$:
\begin{align}
    \mathbf{z}^l_{1:j} &= \varphi^l_{\mathrm{trans}}(\mathbf{z}^{l-1}_{1:j}) \in \mathbb{R}^{j \times d_l}. \nonumber
\end{align}
\item \textbf{Decoding:} The last latent representation $\mathbf{z}^L_{1:j} \in \mathbb{R}^{d_L}$ is usually aggregating the sequences before it is then decoded to obtain the output $y' \in \mathcal{V}$:
\begin{align}
    y' &= \varphi_{\mathrm{dec}}(\mathbf{z}^L_{1:j}) \in \mathcal{V}, \nonumber
\end{align}
\noindent where $\varphi_{\mathrm{dec}}: \mathbb{R}^{d_L} \to \mathcal{V}$ maps from the latent space to a token in the output space.
\item \textbf{Auto-regression:} In this step, the prediction of the decoder is usually appended to the input, and the process repeats from steps 1 to 4 until halting criteria, such as a stop token, are encountered. This auto-regressive process generates the predicted sequence $\hat{y} \in \mathcal{V}^*$.
\end{enumerate}

We view in-context learning as deeply rooted in {\da} and related to domain-invariant learning, since {\llms} seem to learn shared feature representations $\varphi^l_{\mathrm{trans}}(\cdot)$ that generalize across diverse tasks and domains without explicit retraining.
This is because {\llms} are trained on a large corpus of data that includes various prompt-to-prediction pairs from different domains.
This training enables an {\llms} to form domain-invariant representations that generalize across different tasks, even to those they have not specifically been trained for.
In our paper \cite{Dinu:24}, we accredit this to the associations formed between different domains during training based on the instructions and contextual sequences for problem-specific tasks.

\begin{figure}[!ht]
    \centering
    \includegraphics[width=1.0\linewidth]{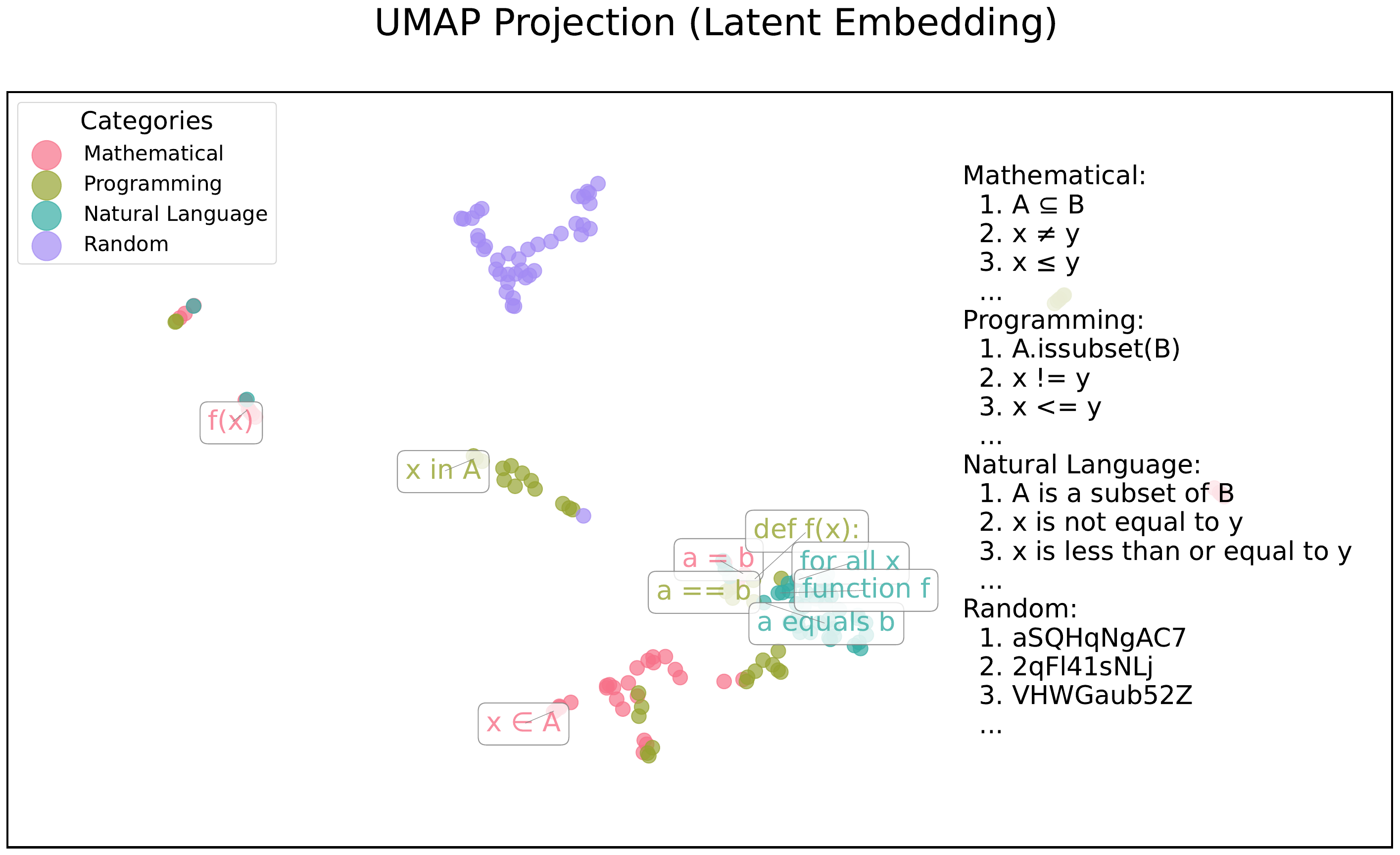}
    \caption{UMAP projection from the latent embeddings of four domains before the classification layer of a GPT-Neo $1.3$ billion parameters {\llm}. The domains are \emph{Mathematical}, \emph{Programming}, \emph{Natural Language} and \emph{Random}. Each domain includes $40$ samples in their respective domain-specific formulation. We see semantically similar concepts cluster together across domains, highlighting the overlap between domains. The \emph{Random} domain is furthest apart, since the embedded sequences do not share semantic overlap with the other domains.}
    \label{fig:latent_umap}
\end{figure}

We draw this conclusion  based on the evidence of analyzing the memorization and learning capabilities of {\llms} \citep{Carlini:21, Min:22, Hartmann:23, VonOswald:23, Duan:24}, polysemanticity and capacity in neural networks \citep{Scherlis:22, Zhao:24}, and semantic feature spaces \citep{Wolfram:23}.
To illustrate, Figure~\ref{fig:latent_umap} presents a UMAP projection \citep{McInnes2018} of the latent space, visualizing the semantic relationships between four different domains based on their embedded representations in the {\llm}.
We show the embedding activations of the last layer of the open-source GPT-Neo $1.3$ billion parameter model \citep{Black:21} for the respective domains.
Here, \emph{domain} represents a specific data distribution of token sequences.
These domains are denoted as \emph{Mathematical}, \emph{Programming}, \emph{Natural Language}, and \emph{Random} domain.
Each dataset includes examples representative of the data distribution of token sequences for the respective domains.

However, we also see an overlap between the first three domains, since they have semantically related features, such as "\emph{a = b}" and "\emph{a equals b}".
This is possible because during training, the model developed a semantic mapping that captures cross-domain correspondences and contextual synonyms.
In-context learning prompts then interpolate between these feature representations by establishing new in-context associations between sequences and, consequently, between data distributions, thereby enforcing domain-invariant predictions.
The power of in-context learning lies in its ability to generalize representations across various tasks and domains by leveraging semantically related features for context-specific adaptations of data distributions.

\paragraph{Structure of In-Context Learning for Domain Adaptation}
{\llms} are usually trained on associated pairs with instructions $i \in \mathcal{I}$, labeled contexts $c \in \mathcal{C}$ sampled w.r.t. to their marginal source and target distribution respectively, and unlabeled query statements $s \in \mathcal{S}$. The contexts include expected labeled outcomes $y \in \Y$ that create a bridge between domains, and help the model understand relationships between sequences of tokens. Specifically, we denote the following elements:
\begin{itemize}
    \item $\mathbf{i} = (i_u)_{u=1}^{m}$ denotes the instruction coming from the marginal distribution $p_\mathcal{I}$, where each $i_u$ represents a token in the instruction sequence.
    \item $\mathbf{C}_p = \{\mathbf{c}_p^v = [(c_{v,t})_{t=1}^{l_{v_1}}, (y_{v,w})_{w=1}^{h_{v_1}}]\}_{v=1}^{k_1}$ represents source domain examples, where $c_{v,t}$ is the $t$-th token of the $v$-th context sequence of length $l_{v_1}$, and $y_{v,w}$ is the $w$-th token of the corresponding label sequence of length $h_{v_1}$, coming from the marginal distribution $p_\mathcal{C}$.
    \item $\mathbf{C}_q = \{\mathbf{c}_q^v = [(c_{v,t})_{t=1}^{l_{v_2}}, (y_{v,w})_{w=1}^{h_{v_2}})]\}_{v=1}^{k_2}$ represents target domain examples, where $c_{v,t}$ is the $t$-th token of the $v$-th context sequence of length $l_{v_2}$, and $y_{v,w}$ is the $w$-th token of the corresponding label sequence of length $h_{v_2}$, coming from the marginal distribution $q_\mathcal{C}$.
    \item $\mathbf{s} = (s_j)_{j=1}^{n}$ is an unlabeled query statement coming from the marginal distribution $q_\mathcal{S}$, where each $s_j$ represents a token in the query sequence.
\end{itemize}

For example, in a sentiment analysis task across different product categories, $\mathbf{i}$ might be "Classify the sentiment as positive or negative:", $\mathbf{c}_p$ could include labeled examples like ("This book was a page-turner!", "very positive") and ("The plot was predictable", "somewhat negative") from the book domain, $\mathbf{c}_q$ could include ("The battery life is impressive", "extremely positive") and ("The screen cracks easily", "very negative") from the electronics domain, and $\mathbf{s}$ could be an unlabeled review "My new book arrived with a crack on its cover" to classify.

We then formulate the input to the model as
$\mathbf{z}^1 = \varphi_{\mathrm{enc}}(\mathbf{i}, \mathbf{c}_p, \mathbf{c}_q, \mathbf{s}; \rho)$, where $\varphi_{\mathrm{enc}} : \mathcal{I}^m \times \mathcal{C}^{k_p} \times \mathcal{C}^{k_q} \times \mathcal{S}^n \to \mathbb{R}^{j \times d}$ represents the encoding and concatenation function based on the embedding model $\rho$ (see in-context learning approach step 1. and 2.), with $k_p$ and $k_q$ the respective lengths for the context variables and $j$ the total sequence length.
The function $\varphi_{\mathrm{enc}}$ utilizes two domains, and outputs a latent embedding for the input of subsequent layers $\varphi^{l}_{\mathrm{trans}}$.
We omit the sequence length indices for brevity of notation.

\paragraph{Problem Statement}
The aim is to form non-trivial in-context associations that enable the model to predict $\hat{y} = y \in \Y$, and formulate in-context learning with the principles of {\uda} that leverages few labeled data pairs from both domains to learn associations and generalize to unlabeled query statements.

By learning from these in-context associations during training, the {\llm} is conditioned to associate concepts that are related, and therefore can learn domain-invariant latent representations.
This process aims to minimize the discrepancy between distributions of source and target components in the latent space.
Specifically, for any layer index $l$, the transformation $\varphi^l_{\mathrm{trans}}$ should map the source component $\mathbf{z}^{l-1}_p$ and the target component $\mathbf{z}^{l-1}_q$ of $\mathbf{z}^{l-1}$ to similar regions in the latent space, such that $\varphi^l_{\mathrm{trans}}(\mathbf{z}^{l-1}_p) \approx \varphi^l_{\mathrm{trans}}(\mathbf{z}^{l-1}_q)$.

We obtain $\mathbf{z}^1_p$ by applying the encoding function to the source context sequences only:
\begin{equation}
\mathbf{z}^1_p = \varphi_{\mathrm{enc}}(\mathbf{0}, \mathbf{c}_p, \mathbf{0}, \mathbf{0}; \rho),
\end{equation}
where $\mathbf{0}$ represents a placeholder (such as a zero vector or a special masking token) used in place of the instruction, target context, and query components. Similarly, for the target component:
\begin{equation}
\mathbf{z}^1_q = \varphi_{\mathrm{enc}}(\mathbf{0}, \mathbf{0}, \mathbf{c}_q, \mathbf{0}; \rho).
\end{equation}
This formulation explicitly shows how we isolate the source and target contexts in the encoding process, allowing us to analyze their representations separately in the subsequent transformations.

\paragraph{Implicit Contextual Associations}
Such associations are often formed implicitly through token sequences that have specific properties attached to them, i.e., the "\texttt{equals}" statement is related to "\texttt{=}" or "\texttt{the same}" which allows for substitutions, and are innate to many domains such as mathematics or programming.
Therefore, the models profit from many such contextual examples, which also enables us to evaluate unconventional queries as shown in the following observations.

Let us examine several examples to illustrate the capabilities of two state-of-the-art language models, Claude~$3.5$~Sonnet and GPT-4 \citep{openai23, Anthropic:24}, both of which were trained using in-context learning.
\begin{example} Consider the following arithmetic query:
\begin{lstlisting}
@ \ding{210} \ding{58} \ding{203} - 2 * \ding{196} = @
\end{lstlisting}
This query, when interpreted by a model, returns a similar prediction as shown below:
\begin{lstlisting}
Let's solve this step-by-step, following the order of operations (PEMDAS: Parentheses, Exponents, Multiplication/Division, Addition/Subtraction):

First, let's identify the numbers:
@\ding{210}@ = 9
@\ding{203}@ = 2
@\ding{196}@ = 5
Now, our equation looks like this:
9 + 2 - 2 * 5 =
We start with multiplication:
2 * 5 = 10
Now our equation is:
9 + 2 - 10 =
We perform addition and subtraction from left to right:
9 + 2 = 11
11 - 10 = 1

Therefore:
@\ding{210} \ding{58} \ding{203} - 2 * \ding{196} = 1@
The final answer is 1.
\end{lstlisting}
The model predicts the answer $1$, which is semantically correct for a human reader. However, it is interesting to observe that the model interprets the symbols \ding{210}, \ding{203}, \ding{196}, and \ding{58} as numbers $9$, $2$, $5$, and the $+$~operator, respectively.
These symbols are entirely different Unicode encodings and are mapped to different embeddings, yet the model successfully associates them with their corresponding mathematical representations.
\end{example}

Let us now explore how this capability extends to other symbols.
\begin{example}
Consider an operation between two weather symbols:
\begin{lstlisting}
@{\Sun} + {\FilledRainCloud} =@
\end{lstlisting}
When evaluated, the models return a similar prediction as follows:
\begin{lstlisting}
@{\Sun} + {\FilledRainCloud} = {\faRainbow}@
The combination of a sun symbol (@{\Sun}@) and a rain symbol (@{\FilledRainCloud}@) typically represents a rainbow (@{\faRainbow}@). This is because rainbows occur in nature when sunlight and rain are present simultaneously ...
\end{lstlisting}
In this case, not only are the symbols {\Sun}, {\FilledRainCloud}, and {\faRainbow} associated with the words sun, rain, and rainbow, respectively, but we also observe that the $+$~and~$=$~operators follow the same semantics as in the arithmetic example above. This showcases a model's ability to blend different domains --- in this case, weather phenomena and basic arithmetic operations --- to produce a meaningful interpretation.
\end{example}

However, it is important to note that not every query is parsable, as demonstrated in the following example.

\begin{example}
Consider a query using the \emph{Braille} alphabet with numbers and the $+$~operator:

\begin{lstlisting}
@1 + \braille{{bb}} = @
\end{lstlisting}
In this case, the Claude model struggles to associate the number $1$ with the two dots to infer a semantic operation and predict a number:
\begin{lstlisting}{language=Text}
To solve this, we need to understand what the symbol @\braille{{bb}}@ represents.
This symbol is a Braille character, but its numerical value isn't universally defined in the same way as the earlier emoji numbers were. Without additional context about how ...
\end{lstlisting}
This example illustrates the limitations of the model's ability to make associations across different symbol systems, especially when the relationship between the symbols is not explicitly defined or commonly used.
\end{example}

Now, to showcase the capabilities of in-context learning, we will add two examples to help the model \emph{understand} how to interpret the query.

\begin{example}
Consider the adapted query using Braille alphabet with numbers and the $+$~operator, and two added in-context examples:

\begin{lstlisting}
2 + @\braille{{ea}}@ = 3
1 + @\braille{{dd}}@ = 5

1 + @\braille{{bb}}@ =
\end{lstlisting}
In this case, both models interpret the context and associate the Unicode encoding with the in-context examples to deduce the correct answer:
\begin{lstlisting}
Let's approach this step-by-step:

From the first equation, we can deduce:
2 + @\braille{{ea}}@ = 3
@\braille{{ea}}@ = 3 - 2 = 1
From the second equation:
1 + @\braille{{dd}}@ = 5
@\braille{{dd}}@ = 5 - 1 = 4
Now, we need to solve:
1 + @\braille{{bb}}@ = ?
We can see a pattern forming. The Braille symbols seem to represent numbers:
@\braille{{ea}}@ represents 1
@\braille{{bb}}@ would likely represent 2
@\braille{{dd}}@ represents 4
Therefore, the equation 1 + @\braille{{bb}}@ would be equivalent to 1 + 2

So, the answer is:
1 + @\braille{{bb}}@ = 3
Would you like me to explain any part of this reasoning in more detail?
\end{lstlisting}
Note that the \braille{{dd}} symbol represents the number $4$, not $3$, since it is the fourth entry in the encoding table of the Braille sequence.
\end{example}
The last example shows that, with only two provided examples, the models are able to associate the domains of numbers with the Braille alphabet and compute the expected prediction (even though the association for the number $4$ for a human reader is non-intuitive).

In practice, these associations are often more subtle, since they occur in permutations of prompts $\mathcal{X} \times \Y$ that have no explicitly separated instruction, context, query statement and label declarations. To illustrate this concept of domain blending within in-context learning, let's examine a final, more complex example.

\begin{example}
We present the models with a prompt that combines two disparate domains: Python programming and Shakespearean literature. The prompt reads: "\emph{Write a simple Python function to check if a file exists in Shakespeare style}". This unusual combination challenges the model to seamlessly integrate knowledge from both disjoint domains.

The resulting prediction demonstrates the model's ability to interpret novel in-context associations without additional context samples:
\begin{lstlisting}[language=Python]
import os

def doth_file_exist(file_path):
  """
  To exist, or not to exist, that is the question:
  Whether 'tis nobler in the mind to suffer
  ...
  And lose the name of action.--Soft you now,
  The fair file!--Nymph, in thy orisons
  Be all my sins remember'd.

  Args:
  file_path (str): The path whose existence is to be verified.

  Returns:
  bool: True if the file exists, False otherwise.
  """
  return os.path.exists(file_path)

file_path = "path/to/your/file.txt"
if doth_file_exist(file_path):
  print("Verily, the file exists!")
else:
  print("Alas, the file does not exist.")
\end{lstlisting}
The model predicts tokens from both the Python programming and Shakespearean literature distributions, showcasing the model's ability to blend seemingly disparate domains, even though such combinations are highly unlikely in real-world scenarios.
\end{example}
In the last example, the instruction is defined within the sequence of \emph{Write a ... in ... style}.
The source and target domain context is present in an abstract form as \emph{Python programming} and \emph{Shakespeare} respectively.
The query statement is implicitly provided during inference, as {\llms} typically employ special tokens to delineate user prompts from model-generated responses.
Although, the models learned during training to represent both domains independently, due to semantic overlap between tokens, at inference both models are able to interpolate between the domain representations and predict output sequences coming from the joint distribution.

\begin{figure}[!ht]
    \centering
    \includegraphics[width=1.0\linewidth]{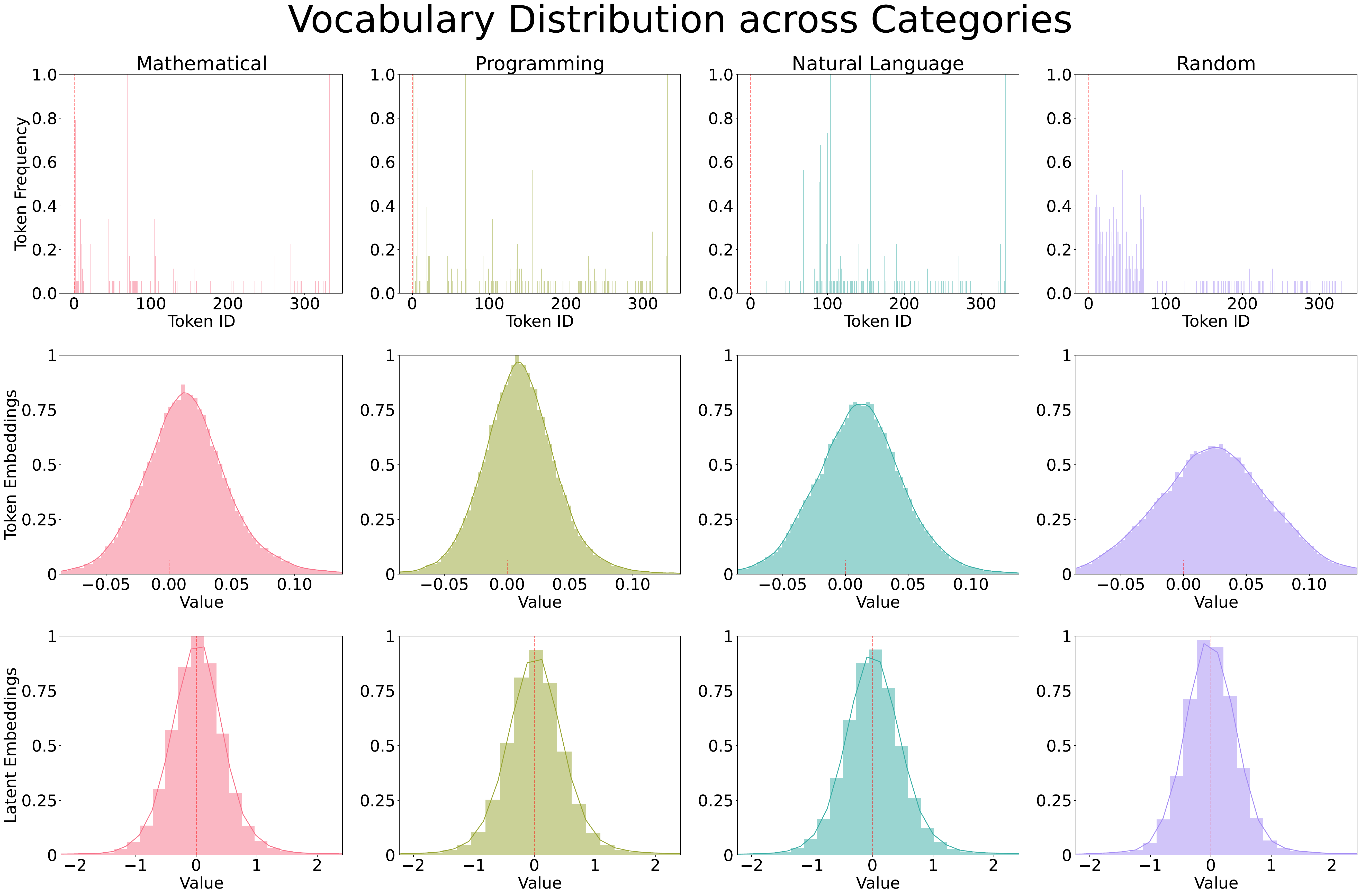}
    \caption{We show the distribution of the tokens across four domains, namely \emph{Mathematical}, \emph{Programming}, \emph{Natural Language} and \emph{Random} domain. The first row shows on the y-axis the $99$-percentile normalized frequency of tokens. The x-axis shows the used unique tokens across all four domains, concatenated in order. The second and third row show the normalized embedding distribution for the encoded tokens after the tokenization and encoding phase and the latent embeddings before the classification layer, respectively.}
    \label{fig:latent_embedding_vocab}
\end{figure}

\paragraph{In-Context Associations and Attention}
To understand the latent representations \(\varphi^l_{\mathrm{trans}}(\cdot)\) in an {\llm} and how they could form domain-invariant features, we also examine the token frequency and embedding distributions, and the self-attention mechanism of modern \emph{Transformer} architectures based on the GPT-Neo model.
In Figure~\ref{fig:latent_embedding_vocab} we show the token frequency per domain and the embeddings after the encoding phase and before the pre-classification latent layer.
The first row shows the absolute frequency of tokens per domain, with the y-axis normalized by the $99$-percentile of the maximum value obtained across all domains.
The x-axis shows only the used unique token IDs across all four domains, concatenated in order, to enable comparison of the token frequencies.
The second and third row show the normalized embedding distribution for the encoded tokens after the tokenization and encoding phase and the latent embeddings before the classification layer, respectively.
Despite the initial differences in token frequencies and distributions in the lower layers, the normalization and transformation processes within the {\llm} architecture result in centered values (around zero) and aligned embedding distributions across domains, setting the premise for achieving  domain-invariant representations as illustrated in the third row.

The self-attention mechanism is an associative memory \citep{Ramsauer:20} that computes a scaled dot-product between query (\(\mathbf{Q}\)) (unrelated to the previously mentioned query statement of the input sequence), key (\(\mathbf{K}\)), and value (\(\mathbf{V}\)) matrices. These matrices are linear projections of the input sequence represented as a matrix \(\mathbf{z} \in \mathbb{R}^{j \times d}\), where \(j\) is the sequence length and \(d\) is the embedding dimension:
\begin{align}
\mathbf{Q} &= \mathbf{z} \mathbf{W}_Q, \quad \mathbf{W}_Q \in \mathbb{R}^{d \times d_k}, \nonumber \\
\mathbf{K} &= \mathbf{z} \mathbf{W}_K, \quad \mathbf{W}_K \in \mathbb{R}^{d \times d_k}, \nonumber \\
\mathbf{V} &= \mathbf{z} \mathbf{W}_V, \quad \mathbf{W}_V \in \mathbb{R}^{d \times d_v},
\end{align}
where \(\mathbf{W}_Q\), \(\mathbf{W}_K\), and \(\mathbf{W}_V\) are learnable weight matrices for the queries, keys, and values, respectively. The matrix multiplication \(\mathbf{z} \mathbf{W}\) projects each token embedding in \(\mathbf{z}\) to a new space, creating the corresponding \(\mathbf{Q}\), \(\mathbf{K}\), or \(\mathbf{V}\) representation. Following \cite{Vaswani:17}, we typically set \(d_k = d_v = d / h = 64\), where \(d_k\) is the dimensionality of the key vectors, and \(h=8\) is the number of attention heads in a multi-head attention setting.

The scaled dot-product is commonly known as attention scores, which we refer to as our transformation function $\varphi_\mathrm{trans}(\cdot)$, and is computed as follows:
\begin{align}
\varphi_\mathrm{trans}(\mathbf{z}) \coloneq \mathrm{Attention}(\mathbf{Q}, \mathbf{K}, \mathbf{V}) &= \mathrm{softmax}\left(\frac{\mathbf{Q} \mathbf{K}^\top}{\sqrt{d_k}}\right) \mathbf{V} ,
\end{align}
where a scaling factor \(\sqrt{d_k}\) is introduced to mitigate the effect of large dot products in high-dimensional spaces. This scaling helps maintain stable gradients during training, particularly for larger values of \(d_k\), and prevents the softmax function from producing small gradients in regions where the input is large in magnitude.

The self-attention function computes a weighted sum of the rows of the value matrix \(\mathbf{V}\) (each row being a value vector), with weights derived from the similarity between corresponding rows of the query matrix \(\mathbf{Q}\) and columns of the key matrix \(\mathbf{K}^\top\).
Here, self-attention refers to the fact that \(\mathbf{Q},\mathbf{K},\mathbf{V}\) all share the same sequence matrix \(\mathbf{z}\). It's worth noting that in other attention settings, different sequence representations can be used for each component.
The resulting attention output \(\varphi_\mathrm{trans}(\cdot)\) associates information from different parts of the input sequence, capturing the dependencies and relationships between tokens.
For networks with multiple layered transformations $\varphi^{l>1}_\mathrm{trans}$, the output is usually normalized before passed on to subsequent layers.
The respective matrices \( \mathbf{W}^l_Q, \mathbf{W}^l_K, \mathbf{W}^l_V \) learn these associations for domain-specific tasks and generalize to the semantics based on the instructions and contextual patterns.
This generalization mechanism then later allows the model to extrapolate semantics across diverse domains when provided with few-shot examples, effectively enabling {\da} through the formation of context-dependent associations in the attention layers.

\paragraph{Domain-Invariance in Language Models}
In the previous sections, we have seen {\llms} able of forming domain-invariant predictions based on our empirical observations, and accredited this to associative patterns learned during training.
We now conclude our analysis by defining domain-invariance in an {\llm} setting, and propose a measurable relationship between semantic similarity of domains and the convergence of their embedded representations.
This relationship forms the basis for future work on quantifying the models' capabilities for cross-domain generalization.

We note that in a general case an {\llm} is a function that maps $f_\theta: \mathcal{P} \to \mathcal{V}^*$ with parameters $\theta \in \Theta$, which produces a sequence of outputs based on the prompt.
This is due to the auto-regressive setting of predicting and sampling the next token and appending it to the input, which is repeated until the halting criteria such as a stop token terminates the sequence.
In this setting, a decoder typically samples according to some sampling strategy (i.e. using a temperature parameter, \emph{beam search}, etc.) to obtain the most likely sequence of token.
However, for simplicity the following definition will focus only on a special case where we omit the auto-regressive process and assume that the context $\mathcal{C} \subset \X \times \mathcal{V}$ considers the output space equal to the vocabulary, and the {\llm} predicts only the next token $\mathcal{V}$.
Furthermore, to define the domain invariance as shown in Example~\ref{ex:fuzzy_equals}, we assume that "\texttt{True}" and "\texttt{False}" are tokens and part of the vocabulary and then formalize the definition as follows:

Let $\mathcal{V}$ be the vocabulary of tokens of an {\llm} and $\mathcal{P} \subset \X \subset \mathcal{V}^*$ be the prompt space.
Define $\mathcal{P} = \mathcal{I}^m \times \mathcal{C}^{k_p} \times \mathcal{C}^{k_q} \times \mathcal{S}^n$ with instruction space $i \in \mathcal{I} \subset \X$, context space $c \in \mathcal{C} \subset \X \times \mathcal{V}$, and query space $s \in \mathcal{S} \subset \X$. Let $p(i, c, s)$ and $q(i, c, s)$ denote the joint distributions for the source and target domains respectively, with $p_\mathcal{C}$ and $q_\mathcal{C}$ being their respective marginal distributions over the context space. The sequence dimensions $m, k_p, k_q, n$ are each $\geq 1$, the sequence length is $j = m + k_p + k_q + n$, and the first latent dimension is $d_1 \geq 1$.

Now, let an {\llm} be a function $f_\theta: \mathcal{P} \to \mathcal{V}$ with parameters $\theta \in \Theta$, with transformation layers $\varphi^l_{\mathrm{trans}} : \mathbb{R}^{j \times d_{l-1}} \to \mathbb{R}^{j \times d_l}$ and dimensions $d_l$ for $1 < l < L$, and the last transformation layer $\varphi^L_{\mathrm{trans}} : \mathbb{R}^{j \times d_{L-1}} \to \mathbb{R}^{d_L}$ before the decoder $\varphi_{\mathrm{dec}} : \mathbb{R}^{d_L} \to \mathcal{V}$, where the output dimension $d_L = |\mathcal{V}| \geq 1$ is equal to the vocabulary size.

\begin{xdefinition}[Domain-Invariant Language Models]
\label{eq:domain_invariance}
We say an {\llm} has domain-invariant properties if, for all pairs of contextual samples with equivalent semantics, represented by $\mathbf{c}_p \equiv \mathbf{c}_q$, where $\mathbf{c}_p \in \mathbf{C}_p$ comes from the marginal distribution $p_\mathcal{C}$ and the semantically equivalent $\mathbf{c}_q \in \mathbf{C}_q$ comes from the marginal distribution $q_\mathcal{C}$, the $\mathrm{argmax}$ (greedy sampling decoder) of their outputs from the last transformation layer $\varphi^L_{\mathrm{trans}}$ is equal.
Formally:
\begin{equation}
\begin{split}
\forall (\z_p^L,\z_q^L) \in \mathbf{Z}^L_p \times \mathbf{Z}^L_q \text{ with } \z_p^L = \varphi^{1:L}(\mathbf{c}_p), \z_q^L = \varphi^{1:L}(\mathbf{c}_q), \text{ and } \mathbf{c}_p \equiv \mathbf{c}_q: \\
\argmax_{i \in \{1,\ldots,|\mathcal{V}|\}} [\mathbf{z}^{L}_p]_{i} = \argmax_{i \in \{1,\ldots,|\mathcal{V}|\}} [\mathbf{z}^{L}_q]_{i}
\end{split},
\end{equation}
where $\z_p^{L}$ and $\z_q^{L}$ are the outputs of the last transformation layer derived from $\textbf{c}_p$ and $\textbf{c}_q$ respectively, with $\mathbf{z}^1_p = \varphi_{\mathrm{enc}}(\mathbf{0}, \mathbf{c}_p, \mathbf{0}, \mathbf{0}; \rho)$ and $\mathbf{z}^1_q = \varphi_{\mathrm{enc}}(\mathbf{0}, \mathbf{0}, \mathbf{c}_q, \mathbf{0}; \rho)$. Here, $\varphi_{\mathrm{enc}} : \mathcal{P} \to \mathbb{R}^{j \times d_1}$ is the embedding and concatenation function, $\rho : \mathcal{V} \to \mathbb{R}^{d_1}$ is an embedding function, and $\mathbf{0}$ denotes the mask tokens for the components.
The latent representations $\mathbf{Z}^l_p = \{\mathbf{z}^{l}_{p,u} = \varphi^{1:l}(\mathbf{c}_{p,u})\}_{u=1}^{h_p}$ and $\mathbf{Z}^l_q = \{\mathbf{z}^{l}_{q,u} = \varphi^{1:l}(\mathbf{c}_{q,u})\}_{u=1}^{h_q}$ come from the respective domains with $h_p \geq 1$ and $h_q \geq 1$ samples, by applying the composition function up to layer index $l$, where $\varphi^{1:l} = \varphi^{l}_{\mathrm{trans}} \circ \varphi^{l-1}_{\mathrm{trans}} \circ \dots \circ \varphi^{2}_{\mathrm{trans}} \circ \varphi^{1}_{\mathrm{trans}} \circ \varphi_{\mathrm{enc}}$.
\end{xdefinition}
The semantic equivalence symbol $\equiv$ refers to selecting consistently equivalent samples from the respective sets, e.g., the number "\texttt{1}" and string "\texttt{one}", which are then pairwise embedded, transformed, and compared.
We then stipulate that for all these pairwise latent representations, their $\mathrm{argmax}$ must be equal, ensuring they map to the same next token.
The determination of semantic equivalence is incumbent upon the modeling process of the distributions and is an integral part of the training process design.
This definition illustrates how the phenomenon of cross-domain generalization in {\llms} can emerge when a sufficiently large and diverse training data corpus is used.
The definition of domain-invariant properties in {\llms} and the semantic equivalence can be viewed as a language-based extension of the \emph{Curry–Howard correspondence}, which originally establishes a direct relationship between computer programs and mathematical proofs \citep{Howard:1980}.

\paragraph{Domain-Invariant Associations}
While Definition~\eqref{eq:domain_invariance} provides a strict criterion for domain invariance, in practice, we often have only a degree of similarity between domain representations, especially in an auto-regressive generative process. To quantify this similarity, we introduce a semantic distance measure that captures the closeness of representations of the model's layers, and provide an $\epsilon$-close domain-invariance approximation.

Let $\mathcal{F}_\Theta$ be the class of {\llms} with $\epsilon$-close domain-invariant associations:
\begin{equation*}
\mathcal{F}_\Theta = \{f_\theta : \mathcal{P} \to \Y \mid \theta \in \Theta, f_\theta \text{ with $\epsilon$-close domain-invariant associations}\}
\end{equation*}

For a model $f_\theta \in \mathcal{F}_\Theta$, we define the semantic distance between domains represented by $\mathbf{C}_p$ and $\mathbf{C}_q$ as:
\begin{align}
d_{\mathrm{sem},\theta}(\textbf{C}_p, \textbf{C}_q) = \min_{\substack{1 < l \leq L}} \max_{\substack{\z_p^{l} \in \mathbf{Z}_p^{l} \\ \z_q^{l} \in \mathbf{Z}_q^{l}}} \|\mathbf{z}^{l}_p - \mathbf{z}^{l}_q\|_2,
\end{align}
where $\z^{l}_p$ and $\z^{l}_q$ are the input representations for the respective layer index $l$ based on $\mathbf{C}_p$ and $\mathbf{C}_q$ following Definition~\eqref{eq:domain_invariance}.

We say that two domains with contextual samples $\textbf{C}_p$ and $\textbf{C}_q$ have $\epsilon$-close domain-invariant associations if $d_{\mathrm{sem},\theta}(\textbf{C}_p, \textbf{C}_q) \leq \epsilon$, thereby setting an upper bound on the difference in representations measured by $d_{\mathrm{sem},\theta}$.
The semantic distance provides an approximation of the degree of domain invariance formed through associations, with smaller distances indicating greater similarity between domain representations. It also captures the idea that as the distance between the transformed representations of samples from two domains decreases, the semantic distance between these domains also decreases. The semantic distance $d_{\mathrm{sem},\theta}$ therefore quantifies this relationship and can be applied to a variety of domains, considering factors such as semantic overlap, structural similarity, and task-specific relevance.

In practice, the $\mathrm{argmax}$ is not applied; instead, more sophisticated auto-regressive sampling techniques are used. Therefore, it suffices that the last layer representations $\mathbf{z}^{L}_p$ and $\mathbf{z}^{L}_q$ are close enough to semantically map to a similar distribution of representations. This is supported by empirical evidence that suggests the existence of semantic \emph{meaning} spaces or a \emph{semantic law of motion}, as described by Stephen Wolfram \citep{Wolfram:23}. The $\epsilon$-close domain-invariant association offers a practical relaxation when perfect invariance is unattainable. This approach provides a quantifiable measure of similarity between domain representations. In an auto-regressive setting, it can leverage the semantic \emph{closeness} between tokens to guide the semantic space towards the desired outcome.

Furthermore, if we assume that all transformation functions $\varphi^\pi_{\mathrm{trans}}$ of the model $f_\theta$ with $\pi > l$ are Lipschitz continuous with Lipschitz constant $K_\pi$, then for all $\mathbf{z}^{L}_p, \mathbf{z}^{L}_q$ based on $\mathbf{C}_p$ and $\mathbf{C}_q$ coming from the respective marginal distributions, the upper bound on the difference in representations in the last layer is given by:
\begin{align}
\begin{split}
\forall (\z_p^L,\z_q^L) \in \mathbf{Z}^L_p \times \mathbf{Z}^L_q: \|\mathbf{z}^{L}_p - \mathbf{z}^{L}_q\|_2 \leq \epsilon \prod_{\pi=l+1}^L K_\pi.
\end{split}
\end{align}
See also the Lipschitz constant of self-attention \citep{Kim:21}.

\paragraph{Hypothesis on Domain-Invariant Associations}
The described setting now provides a framework to formulize the relationship between in-context learning and domain-invariant representation learning in {\llms}.
Based on this, we propose the following hypothesis:

\begin{hypothesis}
\label{hyp:incontext}
For an {\llm} $f_\theta \in \mathcal{F}_\Theta$ with $\epsilon$-close domain-invariant associations the following applies:
Given any pair of samples from domains $\mathbf{C}_p$ (source) and $\mathbf{C}_q$ (target) with their respective latent representations, if their semantic distance $d_{\mathrm{sem},\theta}(\mathbf{C}_p, \mathbf{C}_q) \leq \epsilon$ for some $\epsilon \geq 0$, then:
\begin{equation}
\Delta(q_{Y|X}, q_{f_\theta(X)}) \leq \Delta(p_{Y|X}, p_{f_\theta(X)}) + \delta(\epsilon),
\end{equation}
where:
\begin{itemize}
    \item $X$ is a random variable taking values in the input space $\X$,
    \item $Y$ is a random variable taking values in the output space $\Y$,
    \item $\Delta$ is a discrepancy measure between probability distributions,
    \item $q_{Y|X}$ and $p_{Y|X}$ are the conditional distributions of $Y$ given $X$ for the target and source domains respectively,
    \item $q_{f_\theta(X)}$ and $p_{f_\theta(X)}$ are the distributions of the model's outputs $f_\theta(X)$ in the target and source domains respectively,
    \item $\delta(\epsilon)$ is a function that approaches zero as $\epsilon$ approaches zero.
\end{itemize}
\end{hypothesis}

This hypothesis formalizes the concept that when the semantic distance between samples of a source domain and a target domain is small, the model's discrepancy between its output distribution and the true output distribution on the target domain is bounded by its discrepancy on the source domain plus a small error term. This error term decreases as the semantic distance decreases.

In turn, this also highlights the model's limitations: when a query extrapolates beyond the model's domain capabilities, we obtain meaningless and potentially \emph{hallucinated} results \citep{Huang:23, Huang:24}, as the model attempts to fill knowledge gaps with its existing, but irrelevant, capabilities.
To some degree, this limitation is often obfuscated by blending the generative process with reinforcement learning fine-tuning from human feedback \citep{Ouyang:22, Kaufmann:24}.
However, recent publications have begun to critically examine these aspects \citep{Hicks:24}.
Despite these and other open challenges, the proposed intersection of domain-invariant learning and {\llms}, along with the utilization of their operator-like patterns, offers potential for grounding prompt engineering in a formal framework and advancing {\nesy} approaches, particularly in bridging symbolic reasoning with the adaptive capabilities of neural networks.

\section{A Neuro-Symbolic Perspective on Large Language Models}
\sectionmark{A Neuro-Symbolic Perspective on {\llm}}
Building on the in-context learning capabilities of large language models ({\llms}) and their ability to interpret in-context associations, we can define task-specific prompts that interpret various instructions and contextual patterns, and combine them in a neuro-symbolic ({\nesy}) approach. These prompts leverage the operator-like patterns learned by the model, such as \texttt{equals}, \texttt{is like}, and \texttt{from ... to}, to solve task-specific requirements without the necessity to update model parameters. This {\nesy} approach connects various domains via a language-based interface, utilizing the model's capacity to recognize and apply semantic patterns across domains.

We therefore consider {\llms} suitable as \emph{semantic parsers} \citep{Dinu:24} that can take in various expressions (which are operation descriptions defining how to manipulate strings), interpret them using learned associations, and predict the most probable statistical outcome. These expressions can be composed into complex functions, forming a multi-step generative process that leverages a model's domain-invariant capabilities.

The language-based interface is well-suited for a {\nesy} approach to interface with classical symbolic methods.
We can interchange between solvers and various tools that can verify our predicted outcomes, taking advantage of the model's ability to bridge symbolic and sub-symbolic paradigms. This {\nesy} approach allows for more structured and rule-based approaches, where the model not only predicts statistical associations but also adheres to defined symbolic rules and expressions.
This enhances our ability to verify results by combining the strengths of both neural and symbolic components.

\paragraph{\emph{SymbolicAI} Framework}

In our work \cite{Dinu:24}, we propose a neuro-symbolic approach called \emph{SymbolicAI}. The framework is designed to bridge the gap between symbolic reasoning and neural network capabilities, particularly with the help of {\llms}.

At its core, SymbolicAI performs mappings that take in various symbols and produce symbols as output, where symbols are sequences of tokens drawn from a finite vocabulary $\mathcal{V}$.
These symbols can be of any length and structure, allowing for great flexibility in representation.
The SymbolicAI framework uses these symbols to form more complex expressions, and uses in-context learning to adapt the generative process based on the given instructions, contexts and query statements.
A key concept in this framework is the idea of expressions and their polymorphic contexts. An expression is an operation that defines transformations of symbols to create new symbols. This can be as simple as concatenating two symbols or as complex as applying mathematical operations to them.
When this operation is evaluated, it produces a new symbol based on specific rules. For example, one rule might define how to perform arithmetic operations on symbols representing numbers and strings.
To illustrate this, consider an example where we want to combine the symbols "\texttt{eight}" and "\texttt{8}". We could define a rule that interprets this as addition, resulting in a new symbol "\texttt{16}" as the prediction.

The framework is grounded in formal language theory and grammars, building on the work of linguists and computer scientists Chomsky and Newell \citep{Chomsky:59, Newell:72, Chomsky:03}.
Our {\nesy} approach uses symbolic methods to define sets of rules for relationships between objects, evoke causal structures, and perform formal evaluations. This allows us to define task-specific operations that query {\llms} to guide their predictions and verify the results with other tools and solvers.
An {\llm} can interpret instructions to form associative patterns with the respective data and return a statistically probable answer.
Our framework then combines such operations into more complex expressions.
These expressions guide the generative process of the {\llm} and enable efficient domain-specific adaptations even under continual domain changes without the necessity of parameter updates.
SymbolicAI thus provides a systematic way to represent and manipulate symbolic information and control generative processes.
Our {\nesy} approach is well suited for enabling enhanced reasoning capabilities and building broad~{\ai} systems that verify the results of generated content such as that obtained through {\llms}, while still leveraging their powerful language understanding and generalization abilities. This can lead to more robust and versatile {\ai} systems capable of handling a wider range of tasks, setting the foundation for automated {\ai} workflows. Moreover, the symbolic component of {\nesy} systems can address challenges of computational irreducibility often encountered in complex tasks, allowing for more transparent, interpretable, and discernible results in scenarios where purely neural approaches might lead to unpredictable or difficult-to-verify outcomes.

\paragraph{Computational Graphs, Explainability and Evaluation}

At the core of the proposed framework is its ability to construct complex computational graphs with symbolic and sub-symbolic approaches that use various expressions.
These computational graphs guide the generative process through the in-context learning capabilities of {\llms} via functional methods that build, evaluate and execute prompt instructions and contextual patterns.
These operations form more complex behaviors, adapting the application behavior based on the data and context.

As a key contribution, we propose a quality measure based on the Fréchet distance and its empirical \emph{VERTEX} score that evaluates the performance of computational graphs with respect to reference distributions. Additionally, we propose a benchmark to evaluate various {\llms} suitable for this {\nesy} approach.

\section{List of Publications}\label{sec:list}
In this section, the publications that are directly and more loosely connected to this work are listed,
with short explanations of the publications and my roles in them.
A ``*'' character indicates shared first authorship.

\paragraph{Directly connected publications.}
The following publications are directly connected to this work.
All publications are full conference papers and have been peer-reviewed.
\begin{enumerate}
    \item \citep{Dinu:24}: \textbf{M.-C. Dinu}, C. Leoveanu-Condrei, M. Holzleitner, W. Zellinger, S. Hochreiter. SymbolicAI: A framework for logic-based approaches combining generative models and solvers. In \textit{Third Conference on Lifelong Learning Agents}, PLMR, 2024.
    \item \citep{Dinu:24b} \textbf{M.-C. Dinu}, C. Leoveanu-Condrei, M. Holzleitner, W. Zellinger, S. Hochreiter. SymbolicAI: A framework for logic-based approaches combining generative models and solvers. GenAI4DM Workshop at \textit{The Twelfth International Conference on Learning Representations}, 2024.
    \item \citep{Dinu:23}: \textbf{M.-C. Dinu}, M. Holzleitner, M. Beck, H. D. Nguyen, A. Huber, H. Eghbal-zadeh, B. A. Moser, S. Pereverzyev, S. Hochreiter, W. Zellinger. Addressing parameter choice issues in unsupervised domain adaptation by aggregation. In \textit{International Conference on Learning Representations}, 2023.
    \item \citep{Zellinger:21}: W. Zellinger, N. Shepeleva, \textbf{M.-C. Dinu}, H. Eghbal-zadeh, H. D. Nguyen, B. Nessler, S. Pereverzyev, B. A. Moser. The balancing principle for parameter choice in distance-regularized domain adaptation. In \textit{Advances in Neural Information Processing Systems}, 2021.
\end{enumerate}

\citep{Dinu:23} is the main publication of our work on parameter choice issues for unsupervised domain adaptation.
My main contributions to this work are the implementation of the method, writing of the paper, and development of the benchmark and baseline models.
This includes the design and implementation of the {\iwa} model and its training for the domain adaptation experiments, the realization of the baselines, as well as the investigation of approaches for evaluating the aggregation correlation coefficients of the {\iwa} models.
\citep{Dinu:24} is my most comprehensive publication and focuses on {\nesy} approaches.
My contributions are the conception and implementation of the main approach, as well as the writing of the paper, development of the framework, the proposal and implementation of the VERTEX score, the creation of the benchmark and main parts of the formal derivations.

\paragraph{Loosely connected publications.}
The following publications are loosely connected to this work.
All publications in this section have been peer-reviewed, except for \cite{Patel:24} which is still under review at NeurIPS 2024.
\begin{enumerate}
    \item \citep{Patel:24}: A. Patel, M. Hofmarcher, C. Leoveanu-Condrei, \textbf{M.-C. Dinu}, C. Callison-Burch, S. Hochreiter. Large Language Models Can Self-Improve At Web Agent Tasks. Advances in Neural Information Processing Systems (under review), 2024.
    \item \citep{Dinu:22}: \textbf{M.-C. Dinu}*, M. Hofmarcher*, V. P. Patil, M. Dorfer, P. M. Blies, J. Brandstetter, J. A. Arjona-Medina, S. Hochreiter.
    XAI and Strategy Extraction via Reward Redistribution. International Workshop on Extending Explainable AI Beyond Deep Models and Classifiers. Springer International Publishing, page 177-205, 2022.
    \item \citep{Schweighofer:22}: K. Schweighofer*, A. Radler*, \textbf{M.-C. Dinu}*, M. Hofmarcher, V. Prakash Patil, A. Bitto-Nemling, H. Eghbal-zadeh, S. Hochreiter.
    A dataset perspective on offline reinforcement learning. First
Conference on Lifelong Learning Agents, PMLR, page 470-517, 2022.
    \item \citep{Patil:22}: V. P. Patil*, M. Hofmarcher*, \textbf{M.-C. Dinu}, M. Dorfer, P. M. Blies, J. Brandstetter, J. A. Arjona-Medina, S. Hochreiter.
    Align-rudder: Learning from few demonstrations by reward redistribution. Proceedings of the 39th International Conference on Machine Learning, volume 39, PMLR, page 162:17531-17572, 2022.
    \item \citep{Steinparz:22}: C. A. Steinparz, T. Schmied, F. Paischer, \textbf{M.-C. Dinu}, V. P. Patil, A. Bitto-Nemling, H. Eghbal-zadeh, S. Hochreiter.
    Reactive exploration to cope with non-stationarity in lifelong reinforcement learning. First
Conference on Lifelong Learning Agents, PMLR, page 441-469, 2022.
    \item \citep{Siripurapu:22}: R. Siripurapu, V. P. Patil, K. Schweighofer, \textbf{M.-C. Dinu}, T. Schmied, L. E. F. Diez, M. Holzleitner, H. Eghbal-Zadeh, M. K. Kopp, S. Hochreiter.
    InfODist: Online distillation with Informative rewards improves generalization in Curriculum Learning. Deep Reinforcement Learning Workshop NeurIPS, 2022.
    \item \citep{Schweighofer:21}: K. Schweighofer, M. Hofmarcher, \textbf{M.-C. Dinu}, P. Renz, A. Bitto-Nemling, V. Patil, S. Hochreiter.
    Understanding the effects of dataset characteristics on offline reinforcement learning. Deep RL Workshop NeurIPS, 2021.
    \item \citep{Holzleitner:19}: M. Holzleitner, J. A. Arjona-Medina, \textbf{M.-C. Dinu}, A. Vall, L. Gruber, S. Hochreiter.
    A Two Time-Scale Update Rule Ensuring Convergence of Episodic Reinforcement Learning Algorithms at the Example of RUDDER. NeurIPS Optimization Foundations for Reinforcement Learning Workshop, 2019.
\end{enumerate}

\chapter{Selected Publications}
In this chapter,
our main publications of this cumulative thesis are presented:
\citep{Zellinger:21} in Section~\ref{sec:balancing},
\citep{Dinu:23} in Section~\ref{sec:addressing}, and
\citep{Dinu:24b} in Section~\ref{sec:symbolicai}.
The full list of publications and my contributions are discussed in Section~\ref{sec:list}.

The papers shown in this chapter are shrunk to fit the page layout
and otherwise reprinted as originally published
with permission from the corresponding publishers.
Copyright for the original publications is held by the respective copyright holders.
For the full supplemental materials of~\citep{Zellinger:21},\citep{Dinu:23}~and \citep{Dinu:24} please refer to the respective papers at~\url{proceedings.neurips.cc},  and \url{proceedings.mlr.press}.

\newpage
\section{The balancing principle for parameter choice in distance-regularized domain adaptation}\label{sec:balancing}
\sectionmark{The Balancing principle for parameter choice}
\includepdf[pages={-},scale=0.8,pagecommand={},frame]{"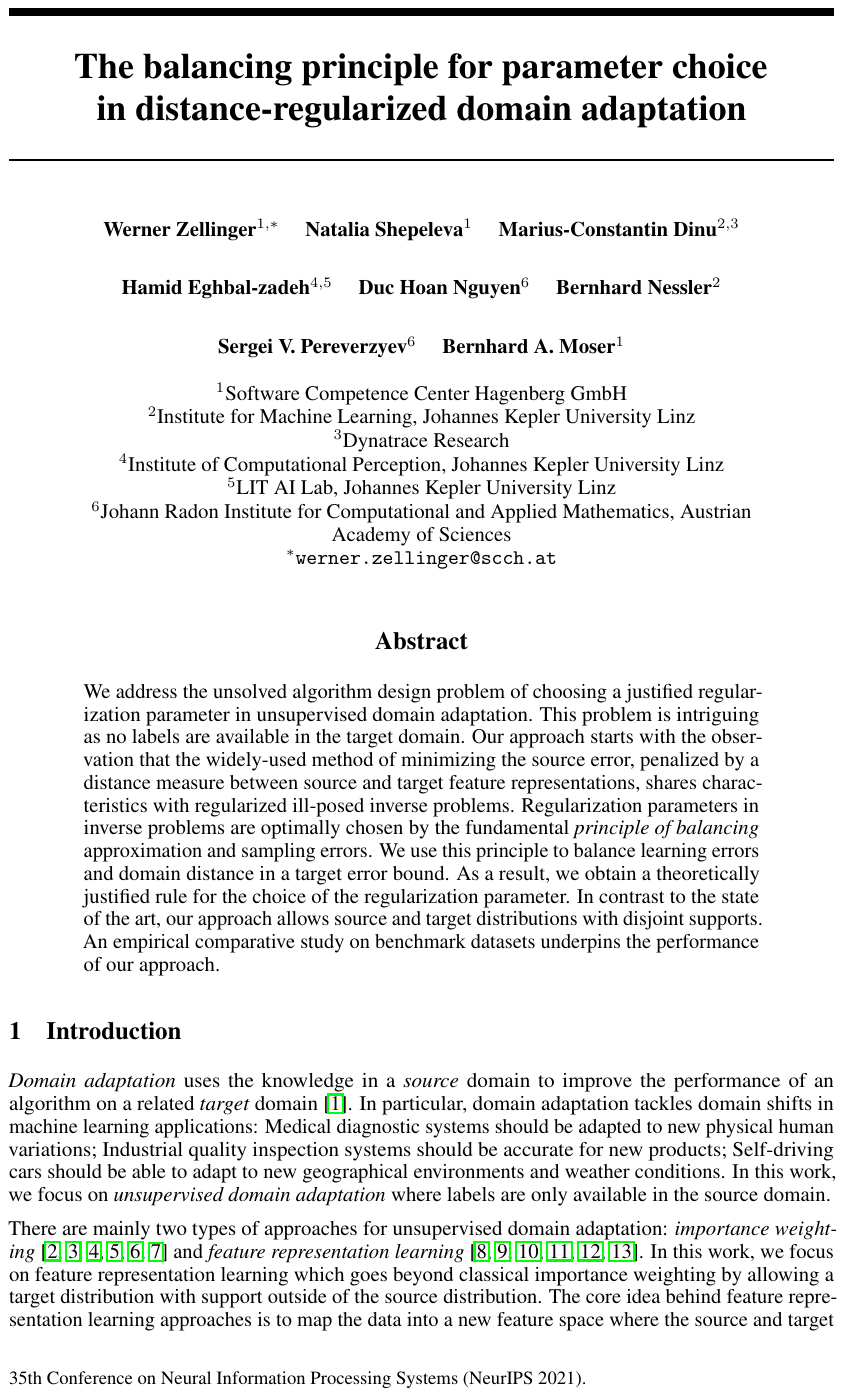"}

\newpage
\section{Addressing Parameter Choice Issues in Unsupervised Domain Adaptation by Aggregation}\label{sec:addressing}
\sectionmark{Addressing parameter choice issues in UDA}
\includepdf[pages={-},scale=0.8,pagecommand={},frame]{"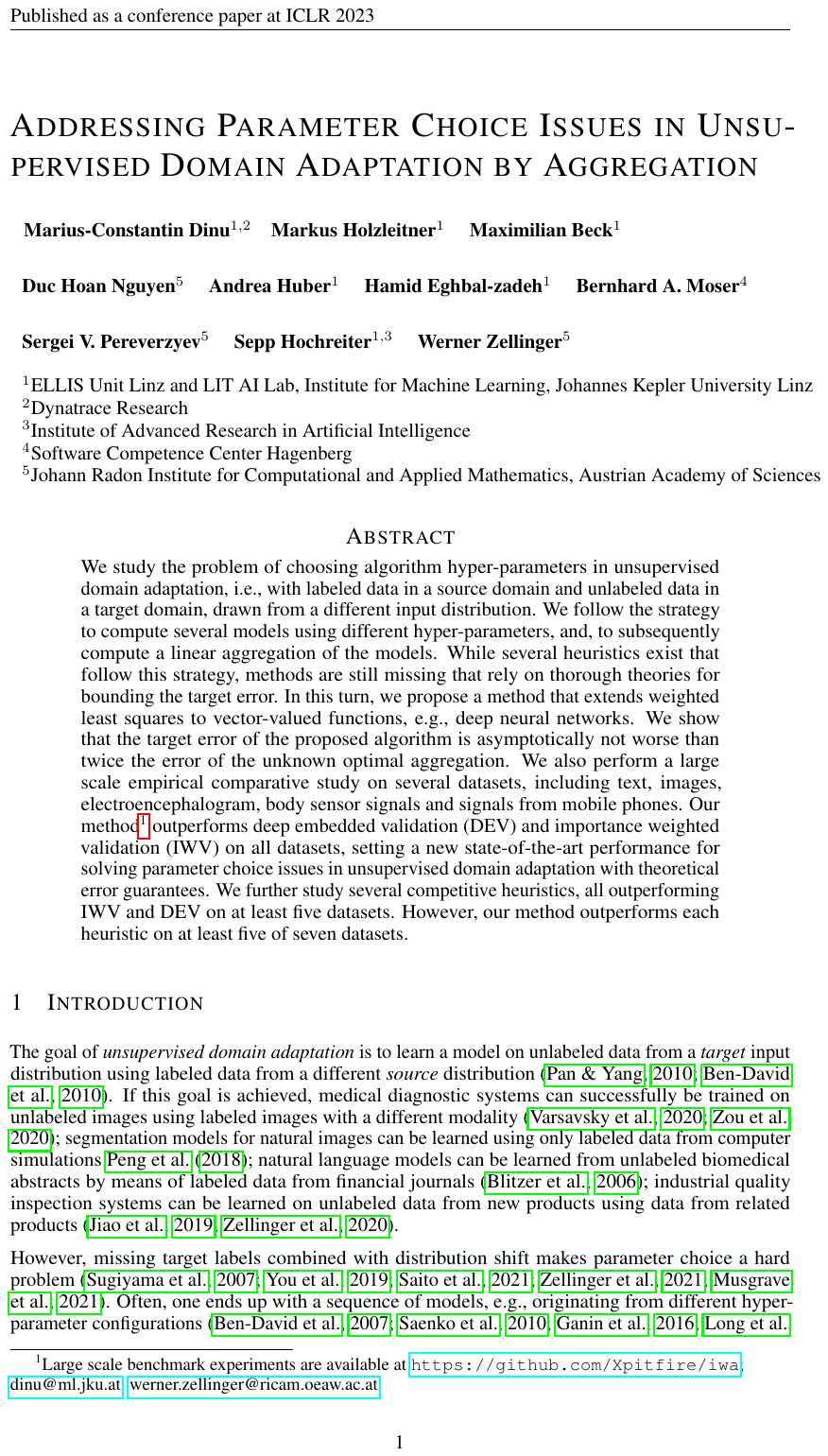"}

\newpage
\section{\emph{SymbolicAI}: A framework for logic-based approaches combining generative models and solvers}\label{sec:symbolicai}
\sectionmark{\emph{SymbolicAI}: Framework}
\includepdf[pages={-},scale=0.8,pagecommand={},frame]{"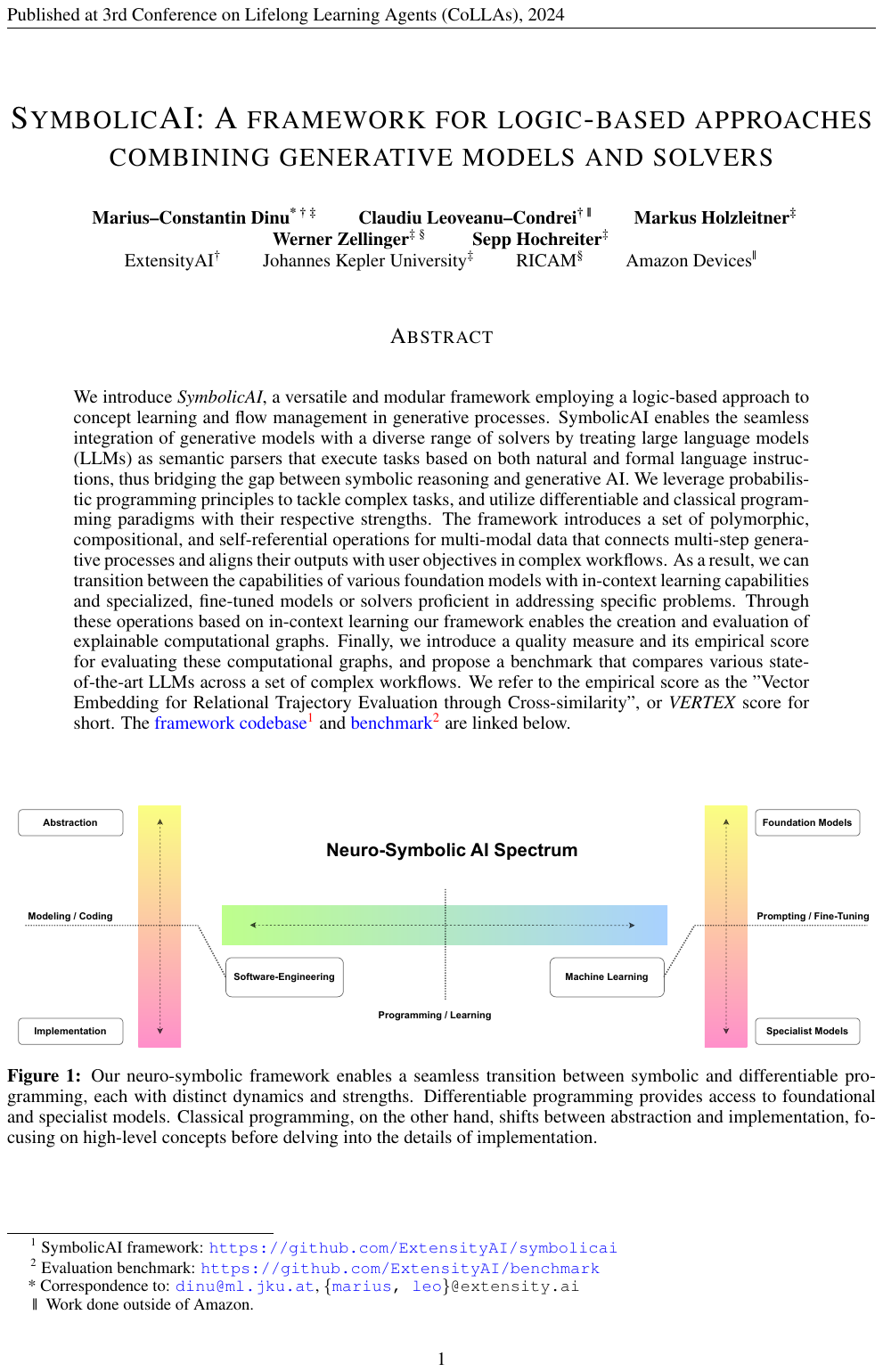"}

\chapter{Conclusion and Outlook}

The main part of this work involves the analysis and development of parameter choice and neuro-symbolic ({\nesy}) approaches for datasets with different characteristics.
It results in the realization of state-of-the-art methods for unsupervised domain adaptation ({\uda}) \citep{Zellinger:21, Dinu:23}, a benchmark for large-scale empirical evaluations on several datasets from different domains \citep{Dinu:23}, and the introduction of a {\nesy} framework for domain-invariant learning without gradient-based updates \citep{Dinu_SymbolicAI:22, Dinu:24, Dinu:24b}.

This work shows a connection between domain-invariant learning and in-context learning.
It also introduced \emph{SymbolicAI}, a logic-based framework that leverages large language models ({\llms}) as semantic parsers to create complex computational graphs, combining symbolic expressions with probabilistic programming paradigms for concept learning and flow management in generative processes, and developed a novel quality measure and benchmark for evaluating these multi-step generative processes across various complex tasks.
We therefore advance the integration between symbolic and sub-symbolic paradigms and open new avenues for {\nesy} architectures that can create in-context associations and behavioral adaptations among abstract representations.

\section{Future Work}

Our future work aims to build upon these foundations by extending our {\uda} benchmark to include additional domains and domain adaptation ({\da}) methods.

The connection between in-context learning to domain-invariant learning supports researcher in approximation theory to perform analysis on the stability and generalizability of learning models across varied domains without retraining.

In continuation of the development of our {\nesy} framework, we plan to explore agentic approaches based on {\llms} and use quantitative measures for capturing the reasoning capabilities of such agents.
By improving the interpretability and robustness of our framework, we aim to ensure its practical applicability in real-world scenarios for autonomous tasks.
Furthermore, we see potential in differentiable computational graphs, synthetic data augmentation of workflows, and the use of the VERTEX score \citep{Dinu:24} as a reward signal for reinforcement learning optimization. We also plan to extend our {\nesy} expression language with more interfaces and workflows.

In summary, by grounding our work in {\da} and {\nesy} approaches, we aim to develop more general, adaptable, interpretable, and scalable {\ai} systems, capable of addressing a wide array of real-world challenges.

\newpage
\bibliography{literature}

\appendix
\renewcommand{\glossarysection}[2][]{}
\chapter{Glossary}
\setlength{\glslistdottedwidth}{0.25\textwidth}  
\setglossarystyle{listdotted}  
\printglossaries

\end{document}